%% file: main.tex
\documentclass[runningheads]{llncs}

\usepackage[year=2026]{eccv}

\usepackage{eccvabbrv}

\usepackage{graphicx}
\usepackage{booktabs}

\usepackage[accsupp]{axessibility}  %

\usepackage[pagebackref,breaklinks,colorlinks,citecolor=eccvblue]{hyperref}
\usepackage{hyperref}

\usepackage{orcidlink}

\usepackage{hyperref}[pagebackref]
\usepackage{url}
\usepackage{float}
\usepackage{xcolor} %
\usepackage{framed} %
\usepackage[most]{tcolorbox} %
\usepackage{listings} %
\usepackage[utf8]{inputenc} %
\usepackage{booktabs}       %
\usepackage{amsfonts}       %
\usepackage{nicefrac}       %
\usepackage{microtype}      %
\usepackage{xcolor}         %
\usepackage{multirow}
\usepackage{graphicx}
\usepackage{amssymb}
\usepackage[normalem]{ulem}
\usepackage{enumitem}
\usepackage{makecell}
\usepackage{subcaption}
\usepackage{wrapfig}
\usepackage{appendix}
\usepackage{fixme}
\fxsetup{
    status=final, %
    author=,      %
    theme=color,
}
\definecolor{lightgray}{rgb}{0.9,0.9,0.9}

\begin{document}

\title{RoadBench: Benchmarking MLLMs on Fine-Grained Spatial Understanding and Reasoning under Urban Road Scenarios}

\titlerunning{RoadBench: Benchmarking MLLMs on Fine-Grained Spatial Reasoning}

\renewcommand{\thefootnote}{\fnsymbol{footnote}}

\newtcolorbox{enhancedpromptbox}{
    enhanced,
    sharp corners,
    boxrule=0.5pt,
    colback=gray!5,
    colframe=gray!50,
    fontupper=\normalfont,
    fonttitle=\bfseries,
    left=10pt,
    right=10pt,
    top=10pt,
    bottom=10pt,
    breakable,
    attach boxed title to top left={xshift=10pt,yshift=-3mm},
    boxed title style={
        sharp corners,
        size=small,
        colback=gray!30,
        colframe=gray!50
    }
}
\lstset{
    basicstyle=\ttfamily\small,
    breaklines=true, %
    backgroundcolor=\color{gray!5},
    frame=none,
    aboveskip=0pt,
    belowskip=0pt
}

\author{Jun Zhang\inst{1} \textsuperscript{*} \and
Xin Zhang\inst{1} \textsuperscript{*} \and
Jie Feng\inst{2} \textsuperscript{\textdagger} \and
Long Chen\inst{3} \and
Junhui Wang\inst{3} \and \\
Zhicheng Liu\inst{3} \and
Depeng Jin\inst{1} \and
Yong Li  \inst{1} \textsuperscript{\textdagger}
}

\authorrunning{J. Zhang, X. Zhang, J. Feng et al.}

\institute{Department of Electronic Engineering, BNRist, Tsinghua University, Beijing, China \and
Zhongguancun Academy \and
Amap, Alibaba Group, Beijing, China \\
\email{fengjie@bza.edu.cn, liyong07@tsinghua.edu.cn}
}

\maketitle

\begingroup
\renewcommand{\thefootnote}{}
\footnotetext{\textsuperscript{*} Equal contribution.}
\footnotetext{\textsuperscript{\dag} Corresponding authors.}
\endgroup

\begin{abstract}

Multimodal large language models (MLLMs) have demonstrated powerful capabilities in general spatial understanding and reasoning.
However, their fine-grained spatial understanding and reasoning capabilities in complex urban scenarios have not received significant attention in the fields of both research and industry.
To fill this gap, we focus primarily on road markings as a typical example of fine-grained spatial elements under urban scenarios, given the essential role of the integrated road traffic network they form within cities.
Around road markings and urban traffic systems, we propose \textbf{RoadBench}, a systematic benchmark that comprehensively evaluates MLLMs' fine-grained spatial understanding and reasoning capabilities using Bird's-Eye View (BEV) and First-Person View (FPV) image inputs.
This benchmark comprises eight tasks consisting of 3,040 strictly manually verified test cases, constructed from 2,137 unique BEV images and 721 unique FPV images collected from five Chinese cities with relatively consistent traffic conventions.
These tasks form a systematic evaluation framework that bridges understanding at local spatial scopes to global reasoning.
They not only test MLLMs' capabilities in recognition, joint understanding, and reasoning but also assess their ability to integrate image information with domain knowledge.  
After evaluating 20 mainstream MLLMs, we confirm that RoadBench is a challenging benchmark for MLLMs while revealing significant shortcomings in existing MLLMs' fine-grained spatial understanding and reasoning capabilities within urban scenarios.
In certain tasks, their performance even falls short of simple rule-based or random selection baselines.
These findings, along with RoadBench itself, will contribute to the comprehensive advancement of spatial understanding capabilities for MLLMs.
The benchmark code is available at \url{https://github.com/tsinghua-fib-lab/RoadBench}, and the supplementary material provides example data, prompts, evaluation scripts, and raw evaluation results.
\end{abstract}

\input{body/1.intro}

\input{body/2.related}
\input{body/3.bench}
\input{body/4.exp}

\input{body/9.conclusion}
\section*{Acknowledgements}
This work was supported in part by the National Key Research and Development Program of China under Grant No. 2024YFC3307603, in part by the National Natural Science Foundation of China under Grant No. 62441229. This work was also sponsored by Tsinghua-Toyota Joint Research Institute Inter-disciplinary Program and the Zhongguancun Academy (Grant No. XTS0074).

\paragraph{Data Ethics and Reproducibility.}

RoadBench is intended solely for academic research and benchmark evaluation. The FPV street-view images are obtained from AMAP with authorization for research use and public release, and all released images have undergone privacy screening, including the removal or blurring of personally identifiable information such as recognizable faces and license plates. We also randomize all sample IDs into UUIDs and convert geographic coordinates into image-level pixel coordinates. The BEV satellite images are based on Google satellite imagery; to support reproducibility while respecting commercial imagery constraints, we provide region-image mappings, geographic coordinates, image filenames, acquisition instructions, construction scripts, prompts, evaluation code, and raw model outputs, enabling users to obtain or reconstruct the images without their own Google API key or paid Google Maps Platform account. Since online map services may update imagery over time, we document the data provenance and provide consistency checks for diagnosing noticeable deviations from the benchmark reference version. RoadBench focuses on five Chinese cities with relatively consistent traffic conventions, and should not be directly used for operational, enforcement, or safety-critical decisions without further ethical review, fairness assessment, and domain-specific validation.

\bibliographystyle{splncs04}
\bibliography{main}
\input{body/supp}

\end{document}

%% file: body/1.intro.tex
\section{Introduction}\label{sec:intro}
\begin{figure}[t]
  \centering
    \includegraphics[width=0.5\textwidth]{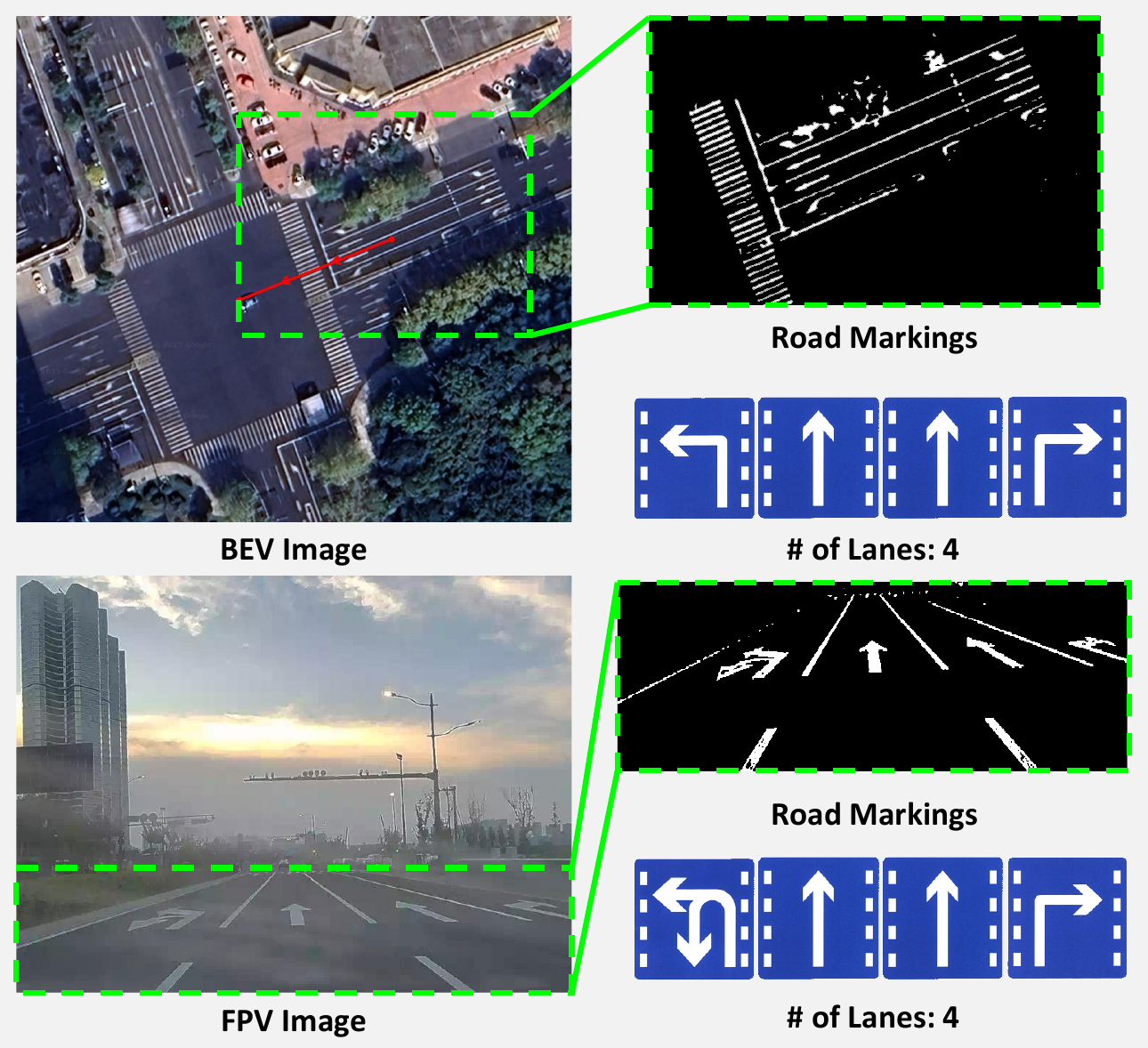}
\caption{Examples of road markings in BEV and FPV images. Road markings such as lane dividers and turning arrows provide fine-grained visual cues for lane counting and lane designation recognition.}
    \label{fig:marking}
\end{figure}

Multimodal large language models (MLLMs) have become a crucial tool for recognizing and understanding the real world due to their powerful combined visual-language comprehension and reasoning capabilities~\cite{achiam2023gpt,yin2024survey,bai2025qwen2,team2025gemma}.
They are progressively replacing specialized models in fields such as satellite image recognition~\cite{zhang2024earthmarker,zhang2026urbanmllm,feng2025urbanllava}, autonomous driving~\cite{cui2024survey,zhang2024interndrive,tian2025drivevlm}, embodied intelligence~\cite{driess2023palm,li2024manipllm}, etc.
Spatial understanding and reasoning in urban environments, as one of the key real-world application scenarios for MLLMs, have also garnered significant attention recently~\cite{roberts2024charting,feng2025citygpt,feng2025urbanllava}.
Various research efforts have released benchmarks~\cite{zhou2025urbench,feng2025citybench,xie2025vlms} for spatial understanding and reasoning tasks in urban environments, evaluating MLLM performance from multiple perspectives.
Existing urban spatial understanding benchmarks provide the foundational data and evaluation criteria necessary to advance MLLM research for real-world applications.

However, we observe that existing benchmarks on urban scenarios mainly focus on whole-image-level understanding or isolated object recognition of specific types, such as GeoQA, landmark recognition, and vehicle detection.
There appears to be a lack of attention to understanding and reasoning about fine-grained spatial elements within urban scenarios.
Recently, cutting-edge research has been exploring how to enhance MLLM's fine-grained understanding and reasoning capabilities in scenarios such as mathematical problems~\cite{zhang2025open} and remote sensing images~\cite{ou2025geopix}.
Nevertheless, studies on MLLM's such capabilities have yet to be applied to urban scenarios, particularly lacking evaluation benchmarks tailored to real-world urban settings.

Under urban scenarios, a representative example of such fine-grained spatial structural elements is the markings painted on city road surfaces shown in Figure~\ref{fig:marking}.
These narrow and long lines and arrows, drawn based on specialized traffic knowledge, collectively form an integrated system that effectively divides and organizes urban space, thereby regulating the movement of pedestrians and vehicles.
The tasks of understanding and reasoning about road markings pose significant challenges to MLLM's capabilities:
\begin{itemize}[leftmargin=1.5em,itemsep=0pt,parsep=0.2em,topsep=0.0em,partopsep=0.0em]
\item The ability to recognize fine-grained structures at a global scale, as road markings are typically thin and extend across the entire image.
\item The joint understanding and reasoning of multiple fine-grained structures, as road markings require overall consideration for accurate semantic recognition.
\item The integration of image information and domain knowledge to generate reasonable responses.
\end{itemize}
We believe that developing MLLM's fine-grained understanding and reasoning capabilities under urban scenarios will not only advance its applications in transportation fields, such as high-definition map auto-generation and end-to-end autonomous driving.
Advancing fine-grained urban road reasoning can also strengthen the general visual-spatial reasoning abilities of MLLMs, as it requires models to understand complex spatial layouts, directional relations, and structured scene semantics.

Therefore, we propose \textbf{RoadBench}, a systematic benchmark primarily designed to evaluate MLLM's understanding and reasoning capabilities regarding fine-grained spatial structural elements under urban scenarios.
RoadBench provides a rich collection of annotated Bird's-Eye View (BEV) and First-Person View (FPV) images, comprising a total of 3,040 test cases across 8 tasks.
These tasks include not only lane counting and lane designation recognition, which are directly based on understanding and reasoning about road markings, but also extend to road network correction and road type classification tasks that rely on joint reasoning involving both road markings and other information within the image.
The test cases comprising RoadBench originate from manually selected images across multiple major cities.
All labels and ground truth data have been labeled and checked by humans to ensure accuracy, with all potentially privacy-sensitive information completely masked and anonymized.
These test cases encompass diverse factors that may impact MLLM understanding, including road and intersection patterns, lighting conditions, seasons, and image resolution. 
Such diversity supports the systematic evaluation of MLLMs' fine-grained spatial understanding and reasoning capabilities.
Based on RoadBench, we conducted a systematic evaluation of 20 mainstream closed-source and open-source MLLMs.
Benchmark results indicate that MLLMs failed to achieve satisfactory outcomes across various tasks, even underperforming against baselines based on random choice or simple rules in certain tasks.
This highlights both the current limitations of MLLMs in achieving fine-grained spatial understanding and reasoning capabilities within urban environments, and underscores RoadBench as a highly challenging benchmark for MLLMs to evaluate such capabilities.

Overall, the main contributions of this paper include the following points:
\begin{itemize}[leftmargin=1.5em,itemsep=0pt,parsep=0.2em,topsep=0.0em,partopsep=0.0em]
    \item We propose a systematic benchmark named \textbf{RoadBench}, designing eight tasks under both BEV and FPV image inputs to comprehensively evaluate MLLMs' understanding and reasoning capabilities regarding fine-grained spatial elements under urban scenarios from local to global scopes.
    \item We collected, processed, and annotated 3,040 test cases as datasets for the eight tasks in RoadBench using data sources such as satellite imagery, online map service providers, and crowd-sourced in-vehicle camera photo databases.
    All test cases underwent rigorous manual verification and annotation to ensure the accuracy of their labels.
    \item We conducted a systematic evaluation of 20 mainstream MLLMs using RoadBench. The results demonstrate that RoadBench is a highly challenging benchmark for MLLMs, while also revealing the limitations of existing MLLMs in fine-grained spatial understanding and reasoning capabilities within urban scenarios.
\end{itemize}

%% file: body/2.related.tex
\section{Related Work}\label{sec:related}

\subsection{MLLM for Spatial Intelligence}

Inspired by the powerful comprehension and complex reasoning capabilities emerging from large language models~\cite{brown2020language,touvron2023llama}, researchers have achieved alignment and fusion between text and image modalities through techniques such as CLIP~\cite{radford2021learning} and BLIP~\cite{li2022blip} to construct multimodal large language models (MLLMs)~\cite{achiam2023gpt,yin2024survey,bai2025qwen2,team2025gemma}.
These MLLMs can simultaneously process textual and visual inputs to perform comprehension and reasoning tasks, enabling them to accomplish a wide range of complex operations~\cite{cui2024survey,xiao2024comprehensive}. 
Additionally, some recent researchers have focused on enhancing MLLM's fine-grained spatial understanding and reasoning capabilities in scenarios such as mathematical problems~\cite{lu2023mathvista,zhang2025open,wei2024slow}, daily images~\cite{azzolini2025cosmos,cheng2024spatialrgpt,guo2024regiongpt,chen2024spatialvlm}, abstract visual puzzles~\cite{wang2024picture,ramakrishnan2024does},  and remote sensing images~\cite{ou2025geopix,hu2025rs,xuan2025dynamicvl}. 
Leveraging the capabilities of MLLM foundation models, some researchers have successfully developed MLLMs specifically tailored for urban scenarios, such as UrbanLLaVA~\cite{feng2025citygpt,feng2025urbanllava}, UrbanMLLM~\cite{zhang2026urbanmllm}, Embodied-R~\cite{zhao2025embodied}, VLDrive~\cite{zhang2025vldrive} and Spatial-LLM~\cite{chen2025spatialllm}. 
However, research on the fine-grained understanding and reasoning capabilities of MLLM in urban scenarios remains relatively rare at present. The lack of benchmarks may be one of the primary factors constraining such work.
Table~\ref{tab:spatial_comparison} compares RoadBench with existing urban-scene and spatial-reasoning benchmarks along key dimensions, including input view, task type, spatial granularity, and road-scene reasoning focus.
\subsection{Spatial Benchmarks for MLLM}

\begin{table*}[t]
\small
\centering
\caption{Comparison of spatial benchmarks for MLLM under real-world urban scenarios.}
\label{tab:spatial_comparison}
\resizebox{\textwidth}{!}{
\begin{tabular}{lccc}
\toprule
\textbf{Benchmark} & \textbf{Minimum Spatial Granularity} & \textbf{Element Inter-dependency} & \textbf{Viewpoint} \\
\midrule
CityBench~\cite{feng2025citybench} & $\bigstar$ (buildings) & $\bigstar$ (isolated elements) & BEV + FPV \\
UrBench~\cite{zhou2025urbench} & $\bigstar$$\bigstar$ (traffic signs) & $\bigstar$$\bigstar$ (cross-view matching) & BEV + FPV \\
CityEQA~\cite{zhao2025cityeqa} & $\bigstar$$\bigstar$ (cars \& buildings) & $\bigstar$ (isolated elements) & 3D Scene (synthetic) \\
DriveBench~\cite{xie2025vlms} & $\bigstar$$\bigstar$ (cars) & $\bigstar$ (isolated elements) & FPV \\
NuPlanQA~\cite{park2025nuplanqa} & $\bigstar$$\bigstar$ (cars \& traffic signs) & $\bigstar$$\bigstar$ (trajectory reasoning) & BEV + FPV \\
MapDR~\cite{chang2025driving} & $\bigstar$$\bigstar$ (lanes \& traffic signs) & $\bigstar$$\bigstar$ (rule-lane matching) & FPV \\
\midrule
\textbf{RoadBench} & \textbf{$\bigstar$$\bigstar$$\bigstar$ (road markings)} & \textbf{$\bigstar$$\bigstar$$\bigstar$ (topological + regulatory logic)} & \textbf{BEV + FPV} \\
\bottomrule
\end{tabular}}
\end{table*}

In recent years, researchers have released extensive benchmarks for MLLMs. For example, MME~\cite{fu2024mmecomprehensiveevaluationbenchmark}, Seed-bench~\cite{li2024seed}, and MMBench~\cite{liu2024mmbench} introduce various perception, cognition, and reasoning tasks to comprehensively evaluate the performance of MLLMs. Some recent researchers have also developed benchmarks using abstract visual puzzles to evaluate the spatial reasoning capabilities of MLLMs, for example, SPACE~\cite{ramakrishnan2024does}, SpatialEval~\cite{wang2024picture}, and Spatial457~\cite{wang2025spatial457}.
VSI-Bench~\cite{yang2025thinking} and MM-Spatial~\cite{daxberger2025mm} evaluate the spatial understanding and reasoning abilities of MLLMs with indoor environment.
In real-world urban scenarios, CityBench~\cite{feng2025citybench} primarily evaluates MLLM's ability to perceive and understand images about cities, as well as its planning and decision-making capabilities.
UrBench~\cite{zhou2025urbench} focuses on evaluating MLLMs in cross-view urban scenarios.
CityEQA~\cite{zhao2025cityeqa} investigates the urban scenarios from the aerial vehicle in a realistic 3D urban simulator.
DriveBench~\cite{xie2025vlms} and NuPlanQA~\cite{park2025nuplanqa} focus on evaluating the reliability of MLLM and multi-view understanding abilities in autonomous driving applications. MapDR~\cite{chang2025driving} is a benchmark dataset designed to evaluate the ability of MLLMs to understand and interpret driving rules through traffic signs.
Overall, as shown in Table~\ref{tab:spatial_comparison}, there is currently no benchmark attentive to the recognition, understanding, or reasoning of fine-grained spatial elements with strong interdependency like road markings under urban scenarios, which limits the evaluation of MLLM capabilities in real world scenarios.

%% file: body/3.bench.tex
\section{RoadBench}\label{sec:bench}

\subsection{Benchmark Overview}

\begin{figure*}[t]
  \centering
    \includegraphics[width=1\textwidth]{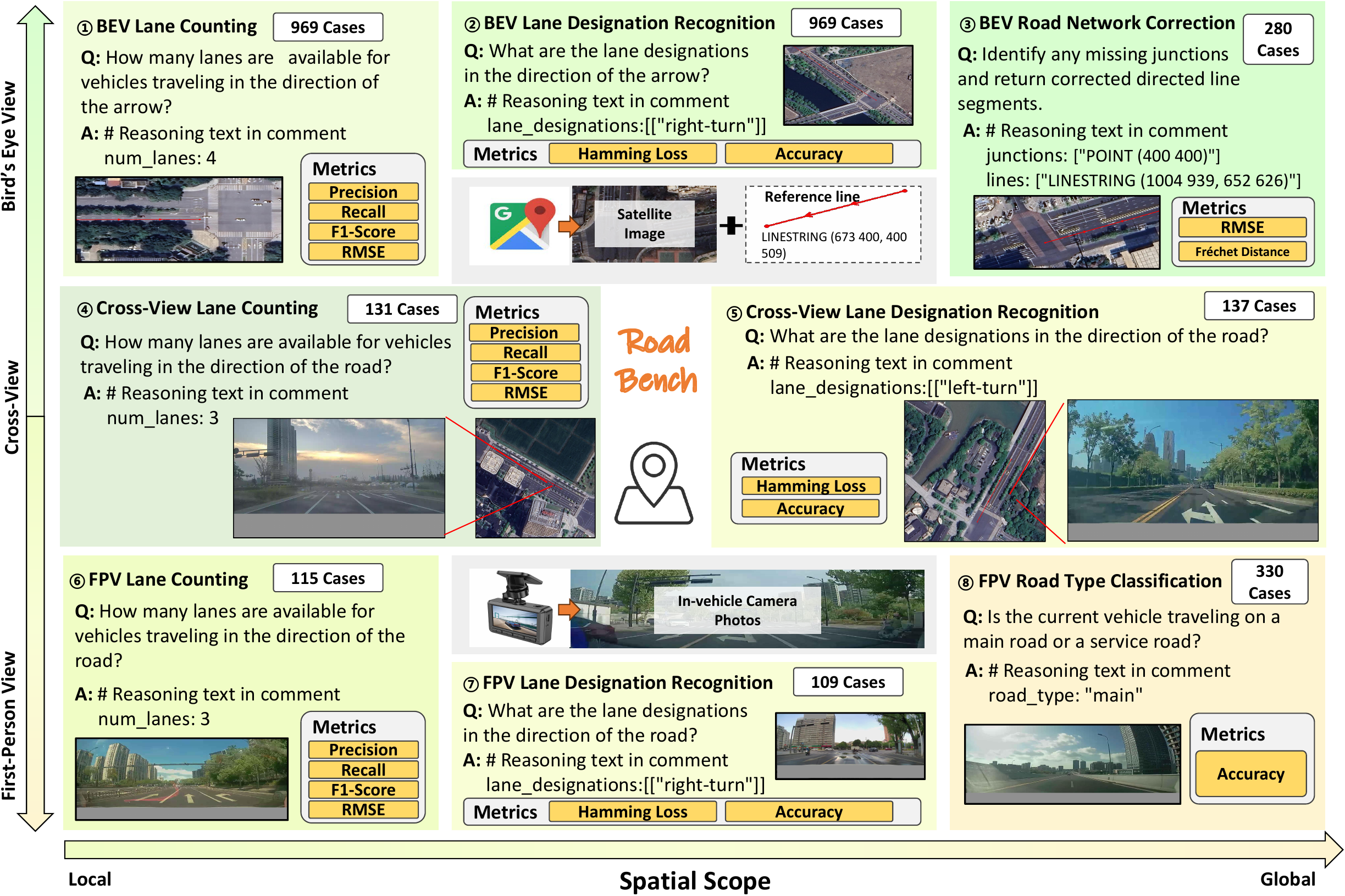}
    \caption{The overview of RoadBench.}
    \label{fig:bench}
\end{figure*}

RoadBench is designed as a benchmark to evaluate the fine-grained spatial understanding and reasoning capabilities of MLLM models in urban scenarios.
As shown in Figure~\ref{fig:bench}, RoadBench contains two types of urban scene images: bird's eye view images derived from satellite imagery and first-person view images captured by in-vehicle cameras.
Based on these two categories of images, we design eight benchmark tasks centered around road markings, which are common fine-grained spatial elements in urban scenarios.
These tasks are organized by the spatial scope required for completion, ranging from local perception to global contextual reasoning. Furthermore, we introduce cross-view tasks that require joint reasoning across BEV and FPV perspectives to evaluate cross-view understanding. Together, they constitute a hierarchical and systematic benchmark spanning fine-grained perception, cross-view synthesis, and global reasoning.
The number of test cases included in each task is also shown in Figure~\ref{fig:bench}.
The entire benchmark comprises a total of 3,040 test cases.
These rigorously hand-curated and processed test cases cover diverse scenarios involving different road and intersection patterns, lighting conditions, seasons, and image resolutions, providing comprehensive coverage of the diversity in urban environments.

\subsection{Benchmark Tasks}

RoadBench comprises eight benchmark tasks, including lane counting and lane designation recognition (utilizing both BEV and FPV imagery), road network correction (BEV-based), and road type classification (FPV-based). Additionally, we include cross-view versions of lane counting and designation recognition to evaluate joint reasoning across perspectives. The following sections detail the setup and evaluation protocols for each task; specific MLLM prompts, representative inputs, and model responses are provided in Appendix \ref{sec:A:prompt}.

\textbf{BEV Lane Counting.}
The lane counting task based on BEV images requires the MLLM to determine the number of lanes contained within the road indicated by the reference line, using the input satellite image and the additional directed reference line.
As shown in the example in Appendix~\ref{sec:A:prompt:task1}, reference lines are drawn as polylines with prominent red arrows on satellite imagery to serve as visual prompts.
The MLLM's output is required to be a text describing the reasoning process along with the number of lanes.
This task aims to conduct a preliminary evaluation of MLLM's ability to follow both visual and textual prompts simultaneously while analyzing and understanding the narrow road markings surrounding reference lines in satellite imagery.
The coordinates of reference lines and actual lane counts in test cases originate from the databases of a tier-1 online map service provider, which also supplies this data for online service delivery.
In terms of evaluation metrics, we employed multi-class classification metrics (Precision, Recall, F1-score) on one hand, while also utilizing RMSE to assess the deviation between MLLM outputs and ground truth.

\textbf{BEV Lane Designation Recognition.}
The lane designation recognition task based on BEV images shares the same MLLM input setup as the BEV lane counting task, including satellite images and reference lines.
In difference, this task further requires MLLM to infer the direction type of each lane by identifying and understanding road surface arrows, among other methods, when the correct number of lanes is known.
Lane direction types include U-turn, left turn, straight ahead, right turn, and their combinations.
This task not only challenges the MLLM's ability to jointly understand a complete set of road markings, but also tests its domain expertise in lane designation.
Ground truth data also comes from the online service database.
Given the combinatorial nature of lane directions, we treat the classification of each lane as an independent multi-label classification problem.
We employ the Hamming Loss as the primary metric and utilize accuracy as a more stringent metric to evaluate whether MLLMs fully comprehend the lane directions.

\textbf{BEV Road Network Correction.}
The road network correction task based on BEV images is both highly challenging and of significant practical value.
This task begins by feeding an inaccurate reference line into MLLMs.
These reference lines often contain errors such as missing actual junctions, which can lead to map services providing incorrect directions to users.
MLLMs are tasked with understanding the concepts of junctions and road segments, then inferring the correct junction locations and road segments based on the diverse information contained within the images to achieve road network correction.
In this task, MLLMs have to recognize not only the image content within the reference line area but also understand broader contextual information such as vehicle orientation and building layout to determine whether any junctions are missing.
The reference lines for this task are sourced from OpenStreetMap~\footnote{\url{https://www.openstreetmap.org/}}, and the actual junction and road segment data also originate from the online map service database.
The evaluation of results is divided into two parts: the accuracy of junction points and the accuracy of road segment polylines.
Since the number of points and polylines returned by MLLMs may not match the ground truth, both evaluations will first match the ground truth with the MLLM output based on the nearest-neighbor principle. Points or polylines that fail to match will be considered as mapped to infinity.
Point matching employs Euclidean distance, and RMSE with a distance upper bound threshold is used as the evaluation metric.
For polylines, given the critical role of direction in road network correction, we employ the Fréchet Distance~\cite{eiter1994computing}, which evaluates directed polyline similarity, as both a distance criterion and a performance metric.
This metric also incorporates a distance upper bound.
Additionally, since image sizes vary, all coordinates are normalized to the range $\left[0, 1\right]$ based on the image length and width before entering the indicator calculation.

\textbf{FPV Lane Counting.}
Similar to the BEV lane counting task, the lane counting task based on FPV images also requires the MLLM to determine the number of lanes from images, sharing identical evaluation metrics and ground truth data sources.
The difference lies in replacing the image perspective with a first-person view captured by an in-vehicle camera.
Simultaneously, the reference lines indicating the target road on the BEV image are removed, requiring MLLMs to understand the spatial relationship between the camera and the surrounding roads to make determinations.

\textbf{FPV Lane Designation Recognition.}
The lane designation recognition task based on FPV images shares identical metrics and ground truth data sources with BEV lane designation recognition task.
Compared to the BEV perspective, which relies entirely on road markings to identify lane direction, MLLMs in the FPV perspective can determine lane direction both through road markings and by confirming overhead signs.
However, FPV images also introduce new challenges for MLLMs. Scenes captured by FPV cameras may include low-light conditions at night or situations where congested traffic obscures road markings. The real-world scenarios reflected in this data will test MLLMs' ability to synthesize information from multiple sources and produce accurate reasoning.

\textbf{FPV Road Type Classification.}
The road type classification task based on FPV images requires MLLMs to determine whether the current vehicle is traveling on the main road or the service road based on the image content.
Unlike other tasks, this task demands that MLLMs go beyond understanding and reasoning about a single category of urban spatial elements, such as road markings or even road networks.
Instead, it requires MLLMs to grasp the semantic information implied by the surrounding environment as a whole and to make inferences based on common sense.
For example, MLLMs can determine whether a road is a service road based on pedestrians or street-side shops in an image, or identify a main road based on enclosed fences and median strips.
Since the dataset construction ensures an equal distribution of test cases between the main and service roads, this constitutes a balanced binary classification task. Thus, the evaluation metric solely employs accuracy.

\textbf{Cross-View Lane Counting.}
Unlike single-view tasks, Cross-View Lane Counting provides geographically paired BEV and FPV imagery at the scene level, requiring the MLLM to perform joint reasoning across disparate perspectives to determine the lane count of a target road. This task exploits the complementary strengths of each modality: the BEV perspective provides a clear topological overview of road geometry, while the FPV perspective offers high-fidelity visual cues, such as lane markings and local traffic context.

To succeed, the model must establish an explicit spatial correspondence between the two views, reconciling top-down structural data with ego-centric visual features. By maintaining the same evaluation metrics and ground-truth sources as the single-view counting tasks, this benchmark provides a controlled and rigorous assessment of the model’s capacity for multi-view information fusion.

\textbf{Cross-View Lane Designation Recognition.}
Building on the cross-view framework, this task requires the MLLM to determine the specific directional designations for each lane. While the BEV perspective defines the underlying structural topology, the FPV perspective contributes essential semantic cues, such as overhead signs and pavement arrows. Success hinges on the model's ability to map these ego-centric visual features onto the global road layout. Consequently, this task serves as a rigorous test of the model’s cross-view semantic alignment and reasoning depth in complex, multi-lane environments.

In summary, the aforementioned eight benchmark tasks systematically evaluate MLLM's ability to understand and reason about fine-grained spatial elements in urban environments across spatial scales from local to global based on images captured from diverse perspectives and scenes.

\subsection{Benchmark Curation}

\begin{figure*}[t]
  \centering
    \includegraphics[width=1\textwidth]{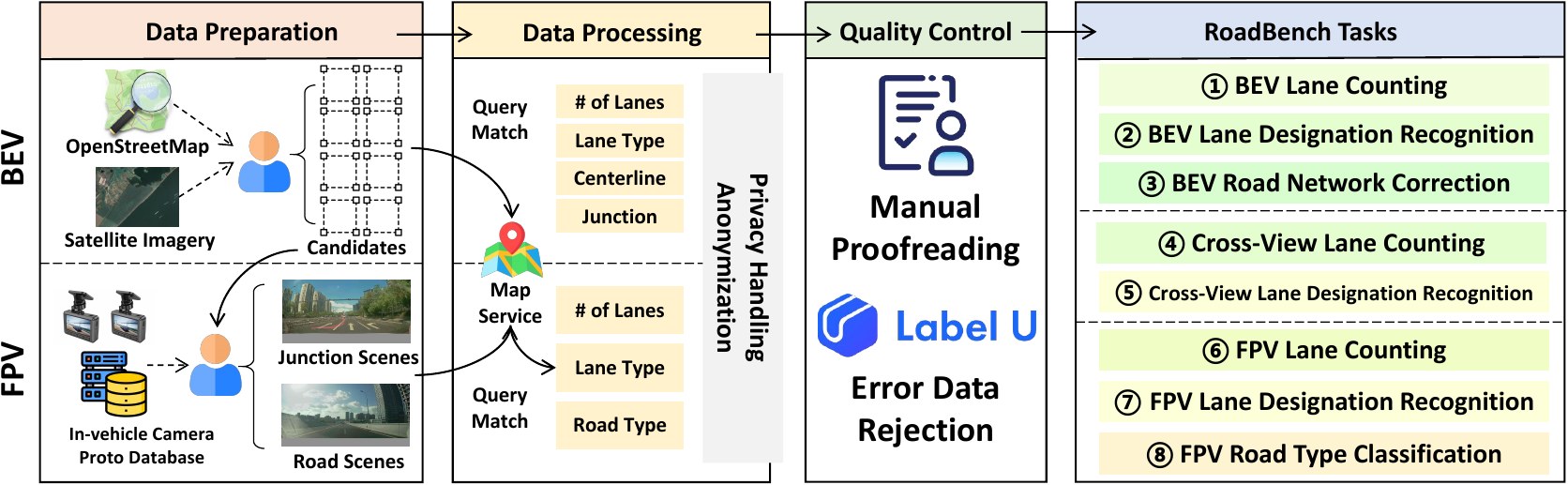}
    \caption{The RoadBench curation pipelines to construct the datasets for the eight benchmark tasks.}
    \label{fig:build}
\end{figure*}

The curation process for RoadBench can be divided into three stages as shown in Figure~\ref{fig:build}: data preparation, data processing, and quality control.
To fully ensure data quality, most steps rely primarily on human expert annotation and rule-driven programmatic automatic matching.

\textbf{Data Preparation.}
For tasks related to BEV images, the primary step in data preparation is determining appropriate spatial scopes based on satellite image resolution and junction patterns.
Specifically, we first download OpenStreetMap data from areas with relatively high satellite image resolution and extract all junctions.
We then manually review the corresponding satellite imagery from Google Maps~\footnote{\url{https://www.google.com/maps}} for each junction to eliminate invalid images that are fake, difficult to identify, or severely obscured, ultimately obtaining the final candidate bounding boxes.
For FPV imagery, we extract valid instances from captured in-vehicle camera photos in the same junction candidates collected for the BEV tasks and road scenes from the same cities to ensure geographic consistency.
For cross-view tasks, BEV-FPV pairs are constructed according to geographic proximity and road-segment correspondence, with mismatched samples manually filtered to ensure reasonable scene-level consistency.
For junction scene images used in lane counting and lane designation recognition, annotators are instructed to include a certain proportion of challenging scenarios, such as nighttime conditions or obscured lane markings, to test the capabilities of MLLMs.
For road scenes, the number of images for main roads and service roads keep consistent to avoid class imbalance issues.

\textbf{Data Processing.}
After completing data preparation, the bounding box data is fed into the database of the tier-1 online map service provider to extract labels such as actual road centerlines, junction locations, the number of lanes, and lane directions within the specified area.
These high-quality datasets are directly applicable to lane-related tasks as ground truth.
Meanwhile, when matched with OpenStreetMap data, they form inaccurate reference lines and ground truth pairs suitable for road network correction tasks.
For FPV images captured at junction and road scenes, the required labels including lane count, lane direction type, and road type are matched from the map provider's database by using the image capture coordinates.
All datasets have undergone anonymization and privacy handling.
All IDs have been randomized into Universally Unique Identifiers (UUIDs).
All location coordinates have been processed into pixel coordinates on the images.
All content in FPV images that could potentially expose the privacy of the photographer or the surrounding environment has been manually obscured using gray masks.

\textbf{Quality Control.}
While the data construction pipeline utilizes automated matching, minor annotation errors may persist due to sensor noise, coordinate misalignments, or algorithmic edge cases. To ensure the highest level of dataset integrity, we conducted a rigorous manual audit of the final test set. Five domain experts performed a comprehensive review and refinement of all samples using the LabelU platform \cite{labelu_2025}. Notably, samples identified as ambiguous or fundamentally erroneous were systematically excluded rather than re-annotated; this conservative filtering strategy ensures the benchmark remains free from subjective human bias and maintains high signal-to-noise quality.

%% file: body/4.exp.tex
\section{Experiments}\label{sec:exp}

\begin{table*}[t]
\centering
\small
\caption{
The comprehensive evaluation results on RoadBench.
The best-performing result is indicated in \textbf{bold}, and the second-best one is indicated with an \underline{underline}.
In the table, F1, HL, and FD are abbreviations for F1-Score, Hamming Loss, and Fréchet Distance, respectively.
For BEV road network correction tasks, the normalized distance upper threshold for the RMSE metric used to evaluate junction point accuracy is 20\%, while the one for the Fréchet Distance metric used to evaluate segment polyline accuracy is 50\%.
}
\label{tab:comprehensive_results}
\setlength{\tabcolsep}{0.5mm}
\resizebox{\textwidth}{!}{
\begin{tabular}{lccccccccccccccc}
\toprule
\multirow{2}{*}{Model} & \multicolumn{2}{c}{\makecell{BEV Lane \\ Counting}} & \multicolumn{2}{c}{\makecell{BEV Lane \\ Designation \\ Recognition}} & \multicolumn{2}{c}{\makecell{BEV Road \\ Network \\ Correction}} & \multicolumn{2}{c}{\makecell{FPV Lane \\ Counting}} & \multicolumn{2}{c}{\makecell{FPV Lane \\ Designation \\ Recognition}} & \multicolumn{1}{c}{\makecell{FPV \\ Road Type\\ Classification }} & \multicolumn{2}{c}{\makecell{Cross-View \\Lane Counting}} & \multicolumn{2}{c}{\makecell{Cross-View Lane \\Designation \\Recognition}} \\
\cmidrule(lr){2-16}
 & F1 $\uparrow$ & RMSE $\downarrow$ & HL $\downarrow$ & Acc. $\uparrow$ & RMSE $\downarrow$ & FD $\downarrow$ & F1 $\uparrow$ & RMSE $\downarrow$ & HL $\downarrow$ & Acc. $\uparrow$ & Acc. $\uparrow$ & F1 $\uparrow$ & RMSE $\downarrow$  & HL $\downarrow$ & Acc. $\uparrow$ \\
\midrule
LLaMA-3.2-11B-Vision & 0.238 & 1.586 & 0.356 & 0.134 & 0.173 & 0.377 & 0.244 & 1.498 & 0.324 & 0.179 & 0.527 & -&-&-&-\\
LLaMA-3.2-90B-Vision & 0.282 & 1.379 & 0.214 & 0.457 & 0.148 & 0.306 & 0.309 & 1.165 & 0.166 & 0.514 & 0.655 &-&-&-&- \\
Qwen2.5-VL-7B & 0.240 & 1.416 & 0.297 & 0.111 & 0.170 & 0.399 & 0.203 & 1.555 & 0.278 & 0.158 & 0.515 &0.227&1.511 & 0.273 & 0.203  \\
Qwen2.5-VL-32B & 0.254 & 1.255 & 0.239 & 0.361 & 0.155 & 0.314 & 0.307 & 1.099 & 0.236 & 0.310 & 0.542 & 0.226 & 1.248 &0.206& 0.408  \\
Qwen2.5-VL-72B & 0.200 & 1.355 & 0.213 & 0.463 & \underline{0.139} & 0.268 & 0.318 & 1.265 & 0.188 & 0.471 & 0.585&0.239&1.384 & 0.183& 0.457  \\
Qwen3-VL-2B & 0.186 & 1.850 & 0.313 & 0.186 & 0.154 & 0.369 & 0.210 & 1.704 & 0.286 & 0.064 & 0.539 & 0.152 & 1.687&0.268&0.100 \\
Qwen3-VL-4B & 0.113 &2.076 & 0.323 & 0.194 & 0.152 & 0.332 & 0.382 & 1.205 & 0.286 & 0.064 & 0.506 & 0.329 & 1.143&0.268&0.100 \\
Qwen3-VL-8B & 0.243 & 1.507 & 0.257 & 0.372 & 0.151 & 0.294 & 0.266 & 1.322 & 0.208 & 0.441 & 0.539 &0.266&1.182&0.217&0.432 \\
Qwen3-VL-32B & 0.276 & 1.299 & 0.188& 0.515 & 0.145 & 0.257 & 0.340 & 1.115 & 0.144 & 0.587 & 0.533 &0.298&1.185&0.175&0.514 \\
Qwen3-VL-30B-A3B & 0.263 & 1.319 & 0.272 & 0.355 & 0.148 & 0.302 & 0.287 & 1.209 & 0.224 & 0.429 & 0.491 &0.279&1.313&0.223&0.408\\
Qwen3-VL-235B-A22B & 0.256 & 1.328 & 0.191 & 0.496 & 0.139 & 0.249 & 0.369 & 1.034 & 0.159 & 0.541 & 0.530 &0.298&1.316&0.178&0.479\\
Gemma-3-12B & 0.207 & 1.214 & 0.256 & 0.341 & 0.156 & 0.289 & 0.200 & 1.212 & 0.263 & 0.343 & 0.548 & 0.196 & 1.201 &0.261&0.347 \\
Gemma-3-27B & 0.279 & 1.334 & 0.214 & 0.399 & 0.146 & \textbf{0.232} & 0.250 & 1.285 & 0.195 & 0.404 & 0.512 & 0.248&1.139 &0.224&0.381\\
Gemini-2.5-Flash & 0.245 & 1.371 & 0.202 & 0.471 & 0.148 & 0.278 & 0.327 & 1.216 & 0.176 & 0.480 & 0.536 & 0.234& 1.182&0.282&0.094\\
Gemini-2.5-Flash-Image & 0.225 & 1.340 & 0.186 & 0.490 & 0.147 & 0.262 & 0.225 & 1.268 & 0.173 & 0.471 & 0.555 & -&-& -&- \\
Gemini-2.5-Pro & {0.302} & 1.252 & 0.172 & 0.524 & 0.140 & 0.289 & \textbf{0.522} & \textbf{0.802} & 0.160 & 0.547 & \textbf{0.748} & \underline{0.365}&1.095& 0.282&0.094\\
GLM-4.5V & 0.289 & \underline{1.183} & 0.175 & 0.535 & 0.171 & 0.268 & 0.322 & 1.071 & 0.145 & 0.571 & 0.570 & 0.329&1.339&0.142&0.557 \\
GPT-5-Nano & 0.197 & 1.529 & 0.171 & 0.540 & \textbf{0.132} & 0.258 & 0.211 & 1.345 & 0.151 & 0.550 & 0.609 & 0.244 &1.384&0.153&0.526 \\
GPT-5-Mini & \textbf{0.355} & \textbf{1.162} & \textbf{0.103} & \textbf{0.662} & 0.141 & \underline{0.235} & 0.362 & 1.009 & 0.137 & {0.584} & 0.588 & 0.344&\underline{1.011}&\textbf{0.121}&\underline{0.588} \\
GPT-5 & \underline{0.305} & 1.350 & {0.159} & {0.579} & 0.143 & 0.261 & \underline{0.498} & \underline{0.923} & \underline{0.122} & \textbf{0.626} & \underline{0.685} & \textbf{0.400}&\textbf{0.878}& \underline{0.124}&\textbf{0.597}\\
\midrule
Rule-based & 0.248 & {1.383} & \underline{0.149} & \underline{0.587} & 0.155 & 0.379 & 0.284 & 1.430 & \textbf{0.127} & \underline{0.605} & 0.458 &0.252&1.395& 0.130&0.588 \\

\midrule
Qwen3-VL-2B(tuned) & 0.319 & 1.204 & 0.232 & 0.338 & 0.114 & 0.190 & 0.387 & 0.987 & 0.103 & 0.678 & 0.830 & 0.421&0.957&0.105&0.668 \\

Qwen3-VL-4B(tuned) & 0.339 & {1.142} & 0.230 & 0.338 & 0.107 & 0.174 & 0.517 & 0.870 & 0.100 & 0.693 & 0.800 & 0.535 &0.829&0.096&0.688\\
Qwen3-VL-8B(tuned) & 0.339 & {1.142} & 0.230 & 0.338 & 0.097 & 0.154 & 0.433 & 0.909 & 0.094 & 0.708 & 0.830 & 0.540&0.824 &0.085&0.728\\

\bottomrule
\end{tabular}
}

\end{table*}

\subsection{Evaluation Settings}

\textbf{Evaluated MLLMs.}
To comprehensively evaluate the performance of MLLMs on RoadBench, we selected mainstream MLLMs released within the last one year and included both open-source and closed-source models with varying parameter counts from different providers.
In the open-source models, we selected the 11B and 90B versions of LLaMA-3.2-Vision, the 7B, 32B, and 72B versions of Qwen2.5-VL~\cite{bai2025qwen2}, the 2B, 4B, 8B, 30B, 32B and 235B versions of Qwen3-VL~\cite{bai2025qwen3}, the 12B and 27B versions of Gemma-3~\cite{team2025gemma}, and GLM-4.5V~\cite{vteam2025glm45vglm41vthinkingversatilemultimodal}.
In the closed-source model selection, we chose Gemini-2.5-Flash, Gemini-2.5-Flash-Image, Gemini-2.5-Pro from Google~\cite{comanici2025gemini}, as well as GPT-5-Nano, GPT-5-Mini, and GPT-5 from OpenAI. Due to temporary model deprecations or API unavailability during the evaluation period, results for certain models are left blank. We also fine-tuned the Qwen3-VL models with 2B, 4B, and 8B parameters on the RoadBench training set.

\textbf{Baselines.}
To aid in understanding the practical applicability of MLLM across these benchmark tasks, we include several rule-based or simple random baselines.
In the two lane counting task, we provided two baselines: always choose two lanes and randomly select two lanes from $\left\{2, 3, 4\right\}$ with a uniform distribution.
For the two lane designation recognition task, we designed a mapping table based on traffic common sense as a baseline. The details can be found in Appendix~\ref{sec:A:result:task2}.
The FPV road type classification task uses uniformly distributed random selection as the baseline.
Due to the complexity of the BEV road network correction task, we directly treat the input reference line as the output road segment polyline, using the start and end points of the reference line as the coordinates of the identified intersection.
The main result shown in Table~\ref{tab:comprehensive_results} will only include the optimal baseline.
Complete experimental results can be found in Appendix~\ref{sec:A:result}.
In addition, since the Quality Control phase only retained data that could be correctly identified by a human, we believe that a human with the related knowledge can perform all tasks completely correctly.
Therefore, no human baseline is provided for comparison.

\textbf{Evaluation Metrics.}
Throughout the entire experiment, the performance of MLLMs across various tasks was comprehensively evaluated using the metrics shown in Figure~\ref{fig:bench}.
The metrics (Precision, Recall, and F1-Score) used to evaluate multi-classification tasks are weighted according to the sample size.
In the BEV road network correction task, the normalized distance upper bound thresholds for RMSE and Fréchet Distance were set to $\left\{10\%, 20\%, 50\%\right\}$ to ultimately select the appropriate threshold.
Due to page size limitations, the main results in Table~\ref{tab:comprehensive_results} report no more than two metrics that best reflect the performance of MLLMs.
The complete results can be found in Appendix~\ref{sec:A:result}.

\textbf{Error Handling.}
To minimize the interference of detectable errors on MLLMs' fine-grained spatial understanding and reasoning evaluation, the benchmark procedure incorporates a series of error-handling mechanisms.
The program detects issues such as failed API calls, empty response values, and incorrect return formats, and re-invokes MLLMs with identical inputs.
This retry process is limited to a maximum of six attempts.
If MLLMs still fail to produce correct results after retries, the outcome is recorded as zero or empty.

\subsection{Main Results}

The comprehensive evaluation results on RoadBench for the selected MLLMs and baselines are listed in Table~\ref{tab:comprehensive_results}.
By analyzing these results, we can find the following conclusions.

\textbf{RoadBench is a highly challenging benchmark for MLLMs.}
Overall, neither the most powerful closed-source models nor open-source models perform sufficiently well on the tasks proposed by RoadBench for evaluating the fine-grained understanding and reasoning capabilities of MLLMs in urban scenarios.
For example, in the BEV lane counting task, the best model GPT-5-Mini achieved an F1-Score of only 0.355, and the RMSE between the predicted lane counts and the ground truth exceeded 1, reaching 1.162.
This indicates that MLLMs fail to effectively and robustly understand the fine-grained spatial structure of road markings in images.
Furthermore, the results of the BEV road network correction task indicate that MLLMs struggle to accurately provide coordinates for corrected intersections and road segments. 
The best model exhibits an RMSE@20\% as high as 0.132 for junction points and an FD@50\% as high as 0.232 for road segment polylines, both falling within the same order of magnitude as the upper threshold set.
These results suggest that current MLLMs are unable to reason based on fine-grained spatial elements and complete tasks requiring global information and also fully demonstrate that RoadBench is a highly challenging benchmark.

\textbf{MLLMs struggle to outperform simple rule-based methods that do not rely on any image inputs.}
Comparing against baselines based on simple rules is a better way to understand the above results than simply interpreting the absolute values of the outcome metrics.
Observing the results of the two lane designation recognition tasks reveals that most MLLMs can not outperform the baseline designed based on traffic domain common sense.
In other scenarios except the FPV road type classification task, some MLLMs also fail to outperform the baseline based on random choice.
By comparing results against simple rules or random selections, we find that MLLMs still have significant room for improvement in fine-grained spatial understanding and the application of domain-specific common sense.

\textbf{MLLMs struggle to provide precise coordinate numbers.}
The BEV road network correction task requires MLLMs to return coordinates for points and polylines based on their understanding and reasoning of the input.
By examining the metrics shown in Table~\ref{tab:task3_results} at the small threshold (10\%), we observe that the results are almost entirely contributed by the distance upper threshold.
This indicates that the points or lines generated by MLLMs shift significantly from the ground truth, reflecting their limitations in spatial understanding and reasoning, or in the accuracy of structured numerical outputs.

\textbf{MLLMs are better at correctly understanding the fine-grained spatial elements contained within FPV images.}
Comparing the results of the same task under FPV and BEV viewpoints, we observe that MLLMs demonstrate significantly superior performance on FPV images compared to BEV images, both in terms of absolute metric values and relative gaps compared to the baseline.
This indicates that larger spatial elements such as road markings and signs in the FPV viewpoint can be better understood by MLLMs.
Conversely, it also reflects the limitations of MLLMs in understanding more granular spatial elements in BEV images.

\textbf{Cross-view reasoning remains challenging for current MLLMs.}
Across the Cross-View Lane Counting and Cross-View Lane Designation tasks, most models exhibit noticeably lower performance compared with single-view tasks. Even strong models such as GPT-5 and Gemini-2.5 show a clear performance drop, indicating that aligning spatial structures between BEV and FPV views remains difficult. Although some models achieve relatively better results, the overall performance gap suggests that robust cross-view spatial understanding is still an open challenge for current multimodal large language models.

\textbf{Fine-tuning substantially improves the performance of small Qwen3-VL models.}
After task-specific tuning, Qwen3-VL-2B, Qwen3-VL-4B, and Qwen3-VL-8B show consistent gains across most RoadBench tasks, with noticeable improvements in lane counting, road type classification, and cross-view reasoning. Among them, Qwen3-VL-8B (tuned) achieves the best overall performance within this group and becomes competitive with several larger open-source models. Nevertheless, a performance gap remains compared with leading closed-source models such as GPT-5 and Gemini-2.5, indicating that model scale and proprietary training pipelines still provide notable advantages.

\textbf{The number of parameters is not a universal solution in RoadBench.}
Although the number of parameters is largely positively correlated with the capabilities of MLLMs, larger models do not necessarily perform better.
For example, GPT-5-Mini outperforms GPT-5 on most tasks.
Qwen2.5-VL-32B also outperformed Qwen2.5-VL-72B in the FPV lane recognition task.
This may be related to the fusion method or training process of the visual and textual modalities within MLLMs.

\textbf{Closed-source models hold certain technical advantages on RoadBench.}
From an overall ranking perspective, in terms of fine-grained spatial understanding capabilities within urban scenarios, closed-source models represented by the GPT-5 series and Gemini-2.5 series  have achieved certain technical advantages over open-source models.
Among open-source models, only GLM-4.5V ranks highly in comprehensive evaluations.

\subsection{Further Analysis}

\textbf{The impact of reference line prompting methods in the BEV tasks.}
Since reference lines serve as the primary source for MLLMs to identify regions of interest in BEV tasks, how their positions are prompted to MLLMs may directly impact the task performance of MLLMs.
We selected the closed-source model GPT-5-Mini and the open-source model GLM-4.5V, which demonstrated better performance in the BEV lane counting task and the BEV lane designation recognition task for further testing.
We design different reference line prompt methods, including text-only prompts, visual-only prompts, and both text and visual prompts.
Visual prompts are categorized into two approaches: using start and end point colors to indicate direction, and employing arrows on the line segments to denote direction.
Both text and visual prompts with arrows are the default settings in the BEV tasks.

Based on the experimental results presented in Table~\ref{tab:glm_prompt_analysis} and Table~\ref{tab:gpt_prompt_analysis}, we observe significant differences in preference for prompt formats among various MLLMs, potentially attributable to variations in the data distribution used during training.
GLM-4.5V shows a stronger tendency to learn from both textual prompts and image prompts with arrows, with image prompts playing a dominant role.
When deprived of image prompts, GLM-4.5V experiences a noticeable performance decline, whereas removing textual prompts results in only minor fluctuations.
GPT-5-Mini shows a stronger preference for prompts that use color to distinguish directions, while different prompting strategies lead to only marginal performance variations in the BEV lane designation recognition task.
These phenomena may indicate that GPT-5-Mini is better equipped to integrate textual and visual information to collaboratively process the entire information flow.

\textbf{The impact of scene environment conditions in the FPV tasks.}
FPV images captured from in-vehicle cameras encompass varying external environmental conditions.
Among these, the most direct factors affecting MLLM understanding and reasoning are adverse lighting conditions and obscured road markings.
Regarding obscured road markings, only images where lane information could be determined through alternative means were retained during manual dataset proofreading.
After additional annotation, 175 test cases involving adverse lighting conditions were identified, alongside 46 test cases featuring obscured road markings.
To analyze the impact of different scenario environments on MLLMs, we selected GLM-4.5V, GPT-5, and Gemini-2.5-Pro which performed well in the main results as case studies.

The experimental results in the FPV lane counting task and the FPV lane designation recognition task are presented in Table~\ref{tab:glm4v_analysis}, Table~\ref{tab:gpt_analysis} and Table~\ref{tab:gemini_analysis}.
Based on the results, we have two primary findings.
First, adverse lighting conditions do indeed degrade the performance of MLLMs, but this impact is significantly alleviated in models with strong image understanding and reasoning capabilities. 
Second, shifting the basis for task completion from obscured road markings to other elements like signage substantially improved the performance of both GPT-5 and Gemini-2.5-Pro.
This phenomenon indicates that MLLMs exhibit significantly weaker capabilities in understanding and reasoning fine-grained spatial elements compared to other capabilities.

%% file: body/9.conclusion.tex
\section{Conclusion}\label{sec:conclusion}

In this paper, we propose a benchmark named RoadBench with eight benchmark tasks and 3,040 test cases for comprehensively evaluating MLLMs' understanding and reasoning of fine-grained spatial elements under urban scenarios based on both BEV and FPV images.
Based on this benchmark, we evaluated 20 mainstream MLLMs.
The results and further analysis indicate that existing MLLMs lack proper fine-grained spatial understanding and reasoning capabilities under urban scenarios.
On certain tasks and metrics, they even fail to outperform baselines based on random selection or simple rules.
These findings indicate the significance of RoadBench while also highlighting the need to enhance MLLM's capabilities in fine-grained spatial understanding and reasoning.
Based on this fact, RoadBench is promising to become the foundational dataset and evaluation framework for advancing research and applications that enhance the fine-grained spatial understanding and reasoning capabilities of MLLM or MLLM-based agents.

%% file: body/supp.tex
\clearpage
\setcounter{page}{1}
\section*{Appendix}
\setcounter{figure}{0}
\setcounter{table}{0}
\renewcommand{\thefigure}{\thesection-\arabic{figure}}
\renewcommand{\thetable}{\thesection-\arabic{table}}

We present the following items in the appendix section:
\begin{enumerate}
    \item The prompts of all the 8 benchmark tasks. (Section~\ref{sec:A:prompt})
    \item Additional experimental setup descriptions and complete results. (Section~\ref{sec:A:result})
    \item Two results supporting further analysis about the impact of reference line prompting methods in the BEV tasks and the impact of scene environment conditions in the FPV tasks. (Section~\ref{sec:A:further})
    \item Analysis of failure cases in all the 8 benchmark tasks. (Section~\ref{sec:A:case})
\end{enumerate}

\input{body/A2.prompt}
\input{body/A3.result}
\input{body/A4.further}
\input{body/A5.case}

%% file: body/A2.prompt.tex
\section{Prompts of Benchmark Tasks}\label{sec:A:prompt}

Examples of prompts for each task in RoadBench and typical MLLM responses are listed below.
It is worth noting that in the prompt design for RoadBench, the requirement for MLLM to return YAML format with a line of comments balances the need for explicit reasoning with the demand for structured output.

\subsection{BEV Lane Counting}\label{sec:A:prompt:task1}

An example image (Figure~\ref{fig:A:task1}) and prompt of the BEV lane counting task are as follows:

\begin{figure}[h]
  \centering
    \includegraphics[width=1\linewidth]{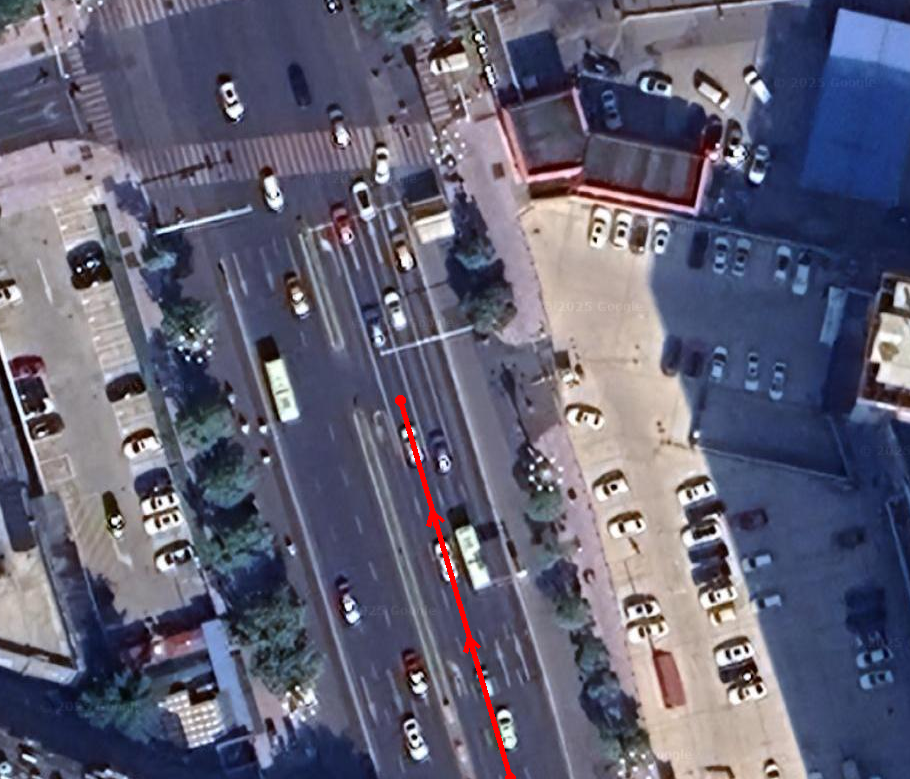}
    \caption{An example image input to MLLM in the BEV lane counting task.}
    \label{fig:A:task1}
\end{figure}

\begin{enhancedpromptbox}
\textbf{Task: Lane Counting for Road Network Analysis}

\textbf{Image Description:}
This is a satellite image of a road network in China, where vehicles drive on the right side of the road. A specific road segment has been marked as a red polyline with arrowheads.

\textbf{Data:}
\begin{lstlisting}
```yaml
# The image size
image:
  width: 910
  height: 779
# The pixel coordinates of the reference centerline
coordinates:
- x: 510.0
  y: 778.0
- x: 400.0
  y: 400.0
```
\end{lstlisting}

\textbf{Question:}
Analyze the marked road segment and determine: How many lanes are available for vehicles traveling in the direction of the arrow?

\textbf{YAML Output Requirements:}
\begin{itemize}[leftmargin=1.5em,itemsep=0pt,parsep=0.2em,topsep=0.0em,partopsep=0.0em]
\item Add ONLY one-line YAML comment explaining your visual analysis and reasoning
\item Extract \texttt{num\_lanes} (integer)
\end{itemize}

\textbf{Output Format:}
\begin{lstlisting}
```yaml
# Explanation: [Describe what you see that led to your determination]
num_lanes: <integer>
```
\end{lstlisting}

Your YAML output:
\end{enhancedpromptbox}

Typical MLLM responses for the image and prompt are as follows:
\begin{enhancedpromptbox}
\textbf{Response from GPT-5:}
\begin{lstlisting}
```yaml
# Explanation: On the right half of the divided roadway (arrow direction), there are three clearly marked lanes separated by dashed white lines between the curb and the central median.
num_lanes: 3
```
\end{lstlisting}

\textbf{Response from Qwen2.5-7B-instruct:}
\begin{lstlisting}
```yaml
Explanation: Observing from top-down view, there appear three distinct parallel lines indicating separate traffic lanes.
num_lanes: 3
```
\end{lstlisting}

\end{enhancedpromptbox}

\subsection{BEV Lane Designation Recognition}

An example image (Figure~\ref{fig:A:task2}) and prompt of the BEV lane designation recognition task are as follows:

\begin{figure}[h]
  \centering
    \includegraphics[width=1\linewidth]{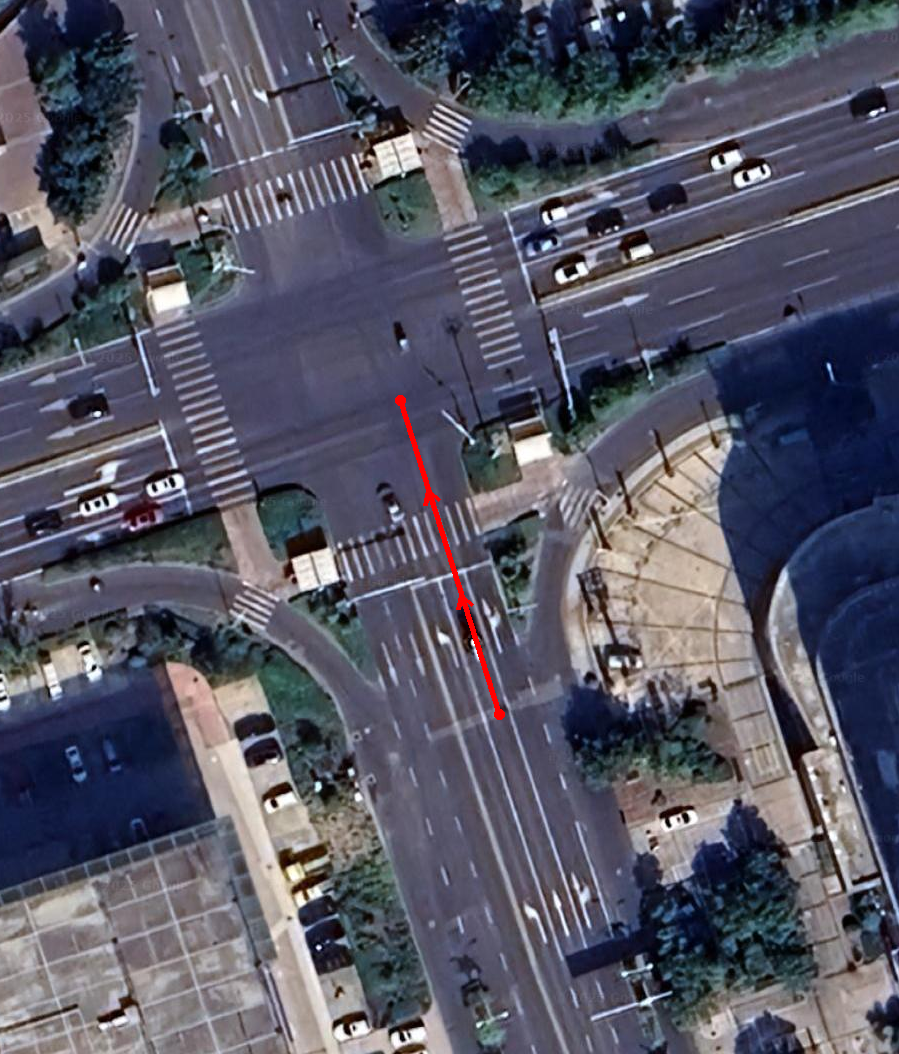}
    \caption{An example image input to MLLM in the BEV lane designation recognition task.}
    \label{fig:A:task2}
\end{figure}

\begin{enhancedpromptbox}
\textbf{Task: Lane Designations Recognition for Road Network Analysis}

\textbf{Image Description:}
This is a satellite image of a road network in China, where vehicles drive on the right side of the road. A specific road segment has been marked as a red polyline with arrowheads.

\textbf{Data:}
\begin{lstlisting}
```yaml
# The number of lanes
num_lanes: 3
# The image size
image:
  width: 899
  height: 1054
# The pixel coordinates of the reference centerline
coordinates:
- x: 499.0
  y: 714.0
- x: 400.0
  y: 400.0
```
\end{lstlisting}

\textbf{Question:}
Analyze the marked road segment and determine: What are the lane designations in the direction of the arrow?

\textbf{Lane Direction Types:}
The available lane direction types include:

(1) U-turn

(2) left-turn

(3) straight

(4) right-turn

\textbf{YAML Output Requirements:}
\begin{itemize}[leftmargin=1.5em,itemsep=0pt,parsep=0.2em,topsep=0.0em,partopsep=0.0em]
    \item Add ONLY one-line YAML comment explaining your visual analysis and reasoning
    \item Return \texttt{lane\_designations} as a list where each item represents the lane direction types for each lane from left to right
    \begin{itemize}
        \item Make sure the number of lane designations is equal to the number of lanes (which is 3)
    \end{itemize}
    \item Each lane can have multiple direction types, so each item in the list should also be a list
    \item Use the English terms: \texttt{"U-turn"}, \texttt{"left-turn"}, \texttt{"straight"}, \texttt{"right-turn"}
\end{itemize}

\textbf{Output Format:}
\begin{lstlisting}
```yaml
# Explanation: [Describe what you see that led to your determination]
lane_designations: [["direction1", "direction2"], ["direction3"], ...]
```
\end{lstlisting}

Your YAML output:
\end{enhancedpromptbox}

Typical MLLM responses for the image and prompt are as follows:
\begin{enhancedpromptbox}
\textbf{Response from GPT-5-mini:}
\begin{lstlisting}
# Explanation: Pavement arrow markings show a dedicated left-turn lane on the left, a straight-only center line, and a rightmost lane marked for both straight and right turns
lane_designations: [["left-turn"], ["straight"], ["straight", "right-turn"]]
\end{lstlisting}

\textbf{Response from Google Gemini-2.5-Flash:}
\begin{lstlisting}
```yaml
# Explanation: From the bottom of the marked segment, the leftmost lane has a straight arrow and a right-turn arrow. The middle lane has a straight arrow. The rightmost lane has a straight arrow.
lane_designations: [["straight", "right-turn"], ["straight"], ["straight"]]
```
\end{lstlisting}

\end{enhancedpromptbox}

\subsection{BEV Road Network Correction}

An example image (Figure~\ref{fig:A:task3}) and prompt of the BEV road network correction task are as follows:

\begin{figure}[h]
  \centering
    \includegraphics[width=1\linewidth]{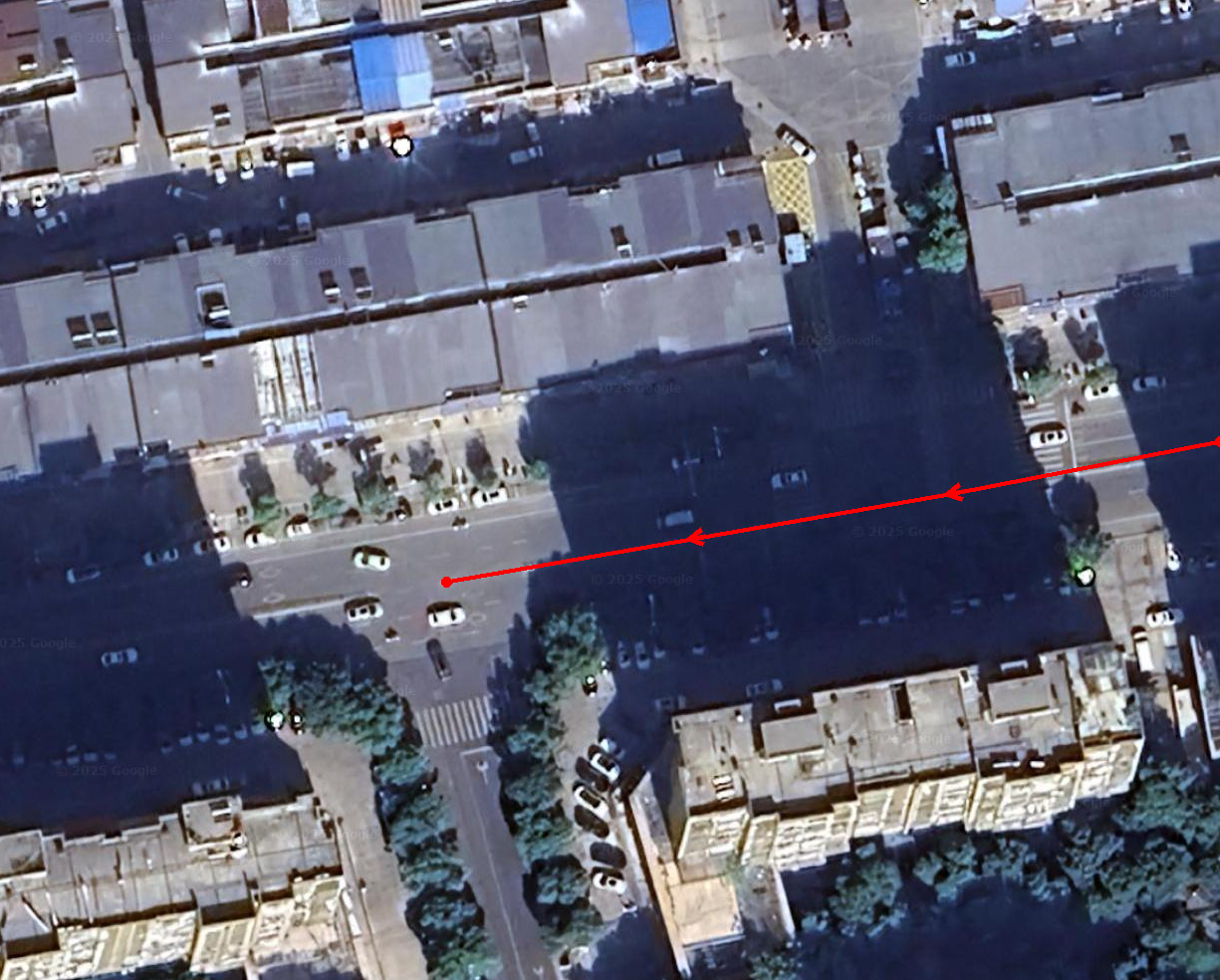}
    \caption{An example image input to MLLM in the BEV road network correction task.}
    \label{fig:A:task3}
\end{figure}

\begin{enhancedpromptbox}
\textbf{Task: Road Network Modification and Junction Identification}

\textbf{Image Description:}
This is a satellite image of a road network in China, where vehicles drive on the right side of the road. A reference line has been marked as a red directed polyline to indicate the path and direction of travel along a road.

\textbf{Data:}
\begin{lstlisting}
```yaml
# The image size
image:
  width: 1225
  height: 984
# The pixel coordinates of the reference centerline
coordinates:
- x: 1225.0
  y: 443.0
- x: 958.0
  y: 495.0
- x: 448.0
  y: 584.0
```
\end{lstlisting}

\textbf{Task Description:}
The given reference line may have missed important junctions such as intersections, highway on/off ramps, U-turn locations, dedicated right-turn lanes, etc., and it may not always be accurately aligned with the true centerline of the road. Your task is to analyze the image and identify:

1. \textbf{Junctions}: Important road intersections, merging points, or decision points that should be marked. Junctions are represented by WKT POINT geometries.

2. \textbf{Line Segments (with Directionality)}: Road segments between junctions that represent the centerlines of the roads, with attention to the direction of travel (from start to end, following the reference line's direction). Line segments are represented by WKT LINESTRING geometries, and their order and orientation should reflect the actual direction of traffic flow along the road.

\textbf{Analysis Requirements:}
\begin{itemize}[leftmargin=1.5em,itemsep=0pt,parsep=0.2em,topsep=0.0em,partopsep=0.0em]
    \item Carefully examine the directed reference line and the actual road network in the image
    \item Identify any missing junctions where the directed reference line should be split (intersections, ramps, etc.)
    \item Return corrected directed line segments that represent the centerlines between identified junctions
    \item Ensure junctions are placed at the center points of intersections/decision points
    \item Ensure line segments follow the actual road centerlines
    \item Do not return junctions or line segments that are not related to the directed reference line
\end{itemize}

\textbf{YAML Output Requirements:}
\begin{itemize}[leftmargin=1.5em,itemsep=0pt,parsep=0.2em,topsep=0.0em,partopsep=0.0em]
    \item Add ONLY one-line YAML comment explaining your analysis and corrections
    \item Return \texttt{junctions} as a list of WKT POINT geometries representing junction center points
    \item Return \texttt{lines} as a list of directed WKT LINESTRING geometries representing road centerlines between junctions
    \item Use pixel coordinates in the format: \texttt{"POINT (x y)"} and \texttt{"LINESTRING (x1 y1, x2 y2, ...)"}
\end{itemize}

\textbf{Output Format:}
\begin{lstlisting}
```yaml
# Explanation: [Describe the junctions and line segments you identified and any corrections made to the directed reference line]
junctions: ["POINT (x1 y1)", "POINT (x2 y2)", ...]
lines: ["LINESTRING (x1 y1, x2 y2)", "LINESTRING (x2 y2, x3 y3)", ...]
```
\end{lstlisting}

Your YAML output:
\end{enhancedpromptbox}

Typical MLLM responses for the image and prompt are as follows:
\begin{enhancedpromptbox}
\textbf{Response from Qwen2.5-VL-72B-instruct:}
\begin{lstlisting}
```yaml
# Explanation: The reference line was extended to include a missed junction at the intersection and corrected to follow the road centerline accurately.
junctions: ["POINT (1225 443)", "POINT (958 495)", "POINT (700 550)", "POINT (448 584)"]
lines: ["LINESTRING (1225 443, 958 495)", "LINESTRING (958 495, 700 550)", "LINESTRING (700 550, 448 584)"]
```
\end{lstlisting}

\textbf{Response from Llama-3.2-11B-vision-instruct:}
\begin{lstlisting}
# Explanation: Corrected and completed the directed reference line, identifying all junctions and centerlines between them, ensuring accurate representation of the road network in China.

junctions: ["POINT (958.0 495.0)", "POINT (448.0 584.0)", "POINT (1225.0 443.0)"]
lines: ["LINESTRING (1225.0 443.0, 958.0 495.0)", "LINESTRING (958.0 495.0, 448.0 584.0)", "LINESTRING (448.0 584.0, 1225.0 443.0)"]
\end{lstlisting}

\end{enhancedpromptbox}

\subsection{FPV Lane Counting}

An example image (Figure~\ref{fig:A:task4}) and prompt of the FPV lane counting task are as follows:

\begin{figure}[h]
  \centering
    \includegraphics[width=1\linewidth]{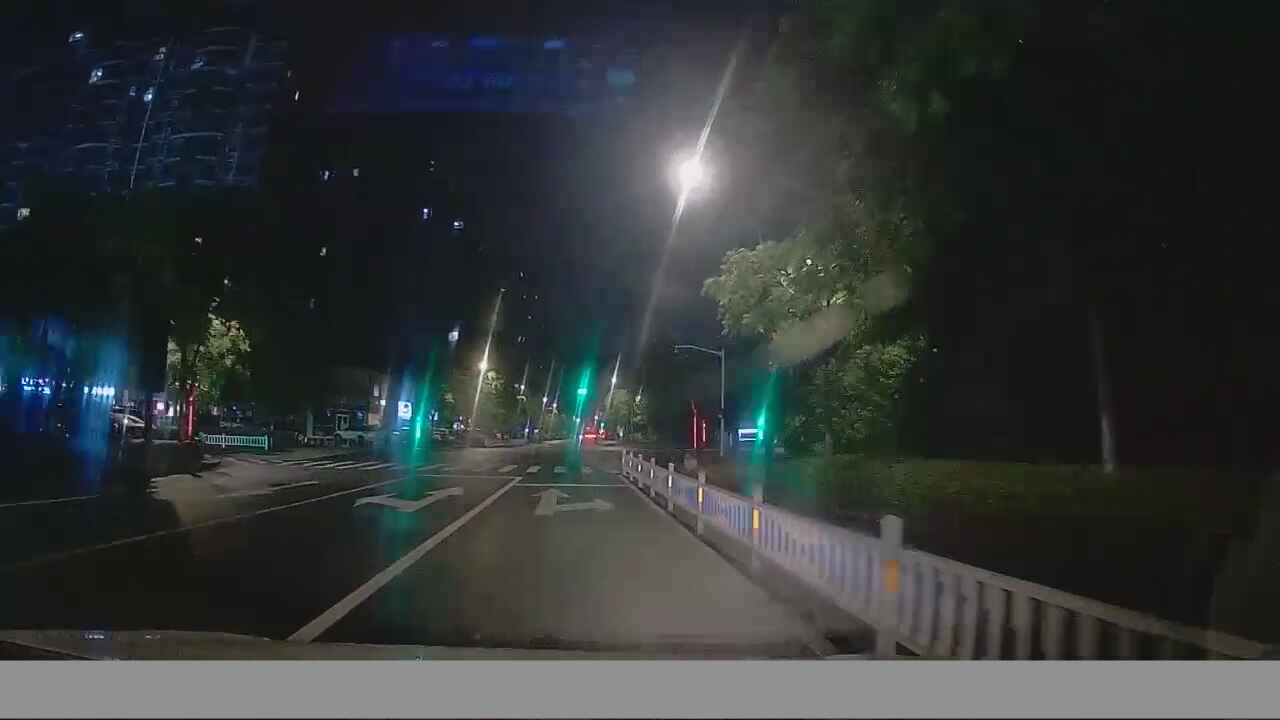}
    \caption{An example image input to MLLM in the FPV lane counting task.}
    \label{fig:A:task4}
\end{figure}

\begin{enhancedpromptbox}
\textbf{Task: Lane Counting for Road Network Analysis}

\textbf{Image Description:}
This is a personal viewpoint image of a road in China, where vehicles drive on the right side of the road.

\textbf{Question:}
Analyze the image and determine: How many lanes are available for vehicles traveling in the direction of the road?

\textbf{YAML Output Requirements:}
\begin{itemize}[leftmargin=1.5em,itemsep=0pt,parsep=0.2em,topsep=0.0em,partopsep=0.0em]
    \item Add ONLY one-line YAML comment explaining your visual analysis and reasoning
    \item Extract \texttt{num\_lanes} (integer)
\end{itemize}

\textbf{Output Format:}
\begin{lstlisting}
```yaml
# Explanation: [Describe what you see that led to your determination]
num_lanes: <integer>
```
\end{lstlisting}

Your YAML output:
\end{enhancedpromptbox}

Typical MLLM responses for the image and prompt are as follows:
\begin{enhancedpromptbox}
\textbf{Response from Llama-3.2-90B-vision-instruct:}
\begin{lstlisting}
```yaml
# The image shows a road with a clear division of lanes through white markings, indicating three distinct lanes for vehicles to travel in the direction shown.
num_lanes: 3
```
\end{lstlisting}

\textbf{Response from Qwen2.5-VL-7B-instruct:}
\begin{lstlisting}
```yaml
# Explanation: The image shows two solid white lines marking separate travel paths.
num_lanes: 2
```
\end{lstlisting}

\end{enhancedpromptbox}

\subsection{FPV Lane Designation Recognition}

An example image (Figure~\ref{fig:A:task5}) and prompt of the FPV lane designation recognition task are as follows:

\begin{figure}[h]
  \centering
    \includegraphics[width=1\linewidth]{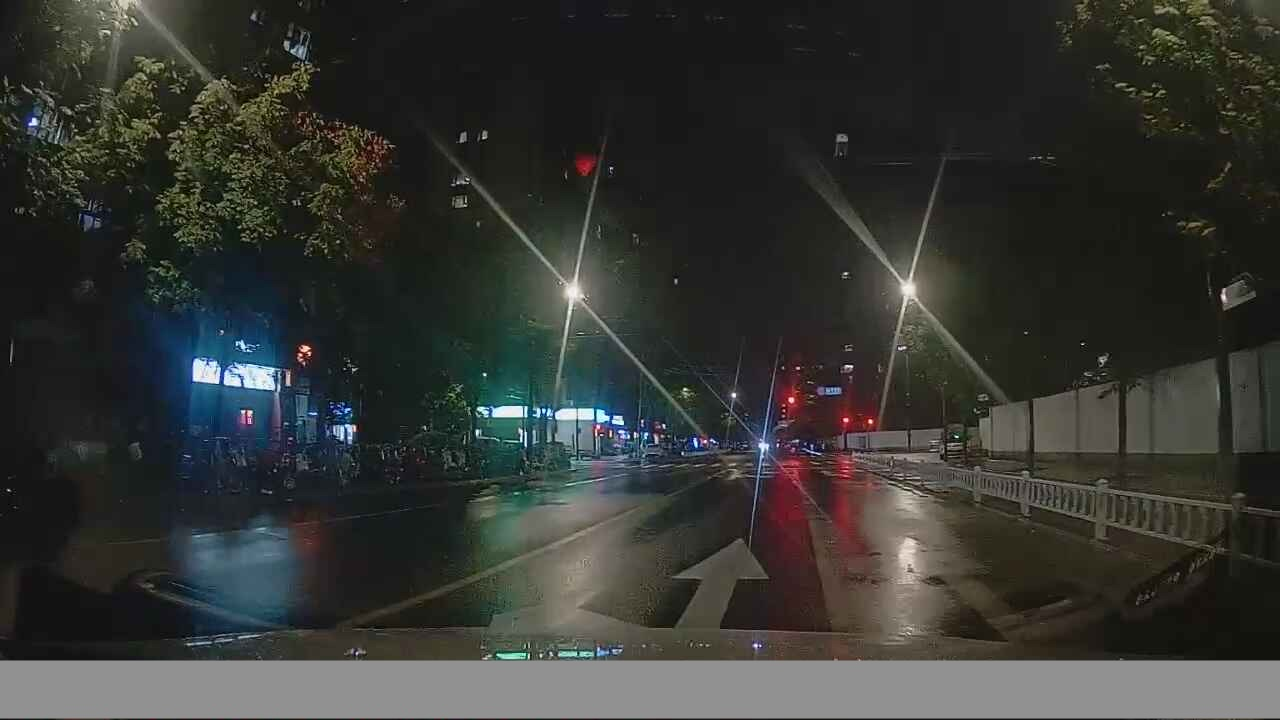}
    \caption{An example image input to MLLM in the FPV lane designation recognition task.}
    \label{fig:A:task5}
\end{figure}

\begin{enhancedpromptbox}
\textbf{Task: Lane Designations Recognition for Road Network Analysis}

\textbf{Image Description:}
This is a personal viewpoint image of a road in China, where vehicles drive on the right side of the road.

\textbf{Data:}
\begin{lstlisting}
```yaml
# The number of lanes
num_lanes: 2
```
\end{lstlisting}

\textbf{Question:}
Analyze the image and determine: What are the lane designations in the direction of the road?

\textbf{Lane Direction Types:}
The available lane direction types include:

(1) U-turn

(2) left-turn

(3) straight

(4) right-turn

(5) variable

\textbf{YAML Output Requirements:}
\begin{itemize}[leftmargin=1.5em,itemsep=0pt,parsep=0.2em,topsep=0.0em,partopsep=0.0em]
    \item Add ONLY one-line YAML comment explaining your visual analysis and reasoning
    \item Return \texttt{lane\_designations} as a list where each item represents the lane direction types for each lane from left to right
    \begin{itemize}
        \item Make sure the number of lane designations is equal to the number of lanes (which is 2)
    \end{itemize}
    \item Each lane can have multiple direction types, so each item in the list should also be a list
    \item Use the English terms: \texttt{"U-turn"}, \texttt{"left-turn"}, \texttt{"straight"}, \texttt{"right-turn"}, \texttt{"variable"}
\end{itemize}

\textbf{Output Format:}
\begin{lstlisting}
```yaml
# Explanation: [Describe what you see that led to your determination]
lane_designations: [["direction1", "direction2"], ["direction3"], ...]
```
\end{lstlisting}

Your YAML output:
\end{enhancedpromptbox}

Typical MLLM responses for the image and prompt are as follows:
\begin{enhancedpromptbox}
\textbf{Response from GLM-4.5V:}
\begin{lstlisting}
```yaml
# Explanation: The image shows a road with two lanes; the visible lane marking indicates a straight direction, and there are no other directional markings (like left-turn or right-turn arrows) for either lane, so both lanes are designated for straight travel.
lane_designations: [["straight"], ["straight"]]
```
\end{lstlisting}

\textbf{Response from GPT-5-nano:}
\begin{lstlisting}
# Explanation: The two lanes ahead have forward (straight) arrow markings with no visible left/right/U-turn arrows, indicating straight travel for both lanes
lane_designations: [["straight"], ["straight"]]
\end{lstlisting}

\end{enhancedpromptbox}

\subsection{FPV Road Type Classification}

An example image (Figure~\ref{fig:A:task6}) and prompt of the FPV road type classification task are as follows:

\begin{figure}[h]
  \centering
    \includegraphics[width=1\linewidth]{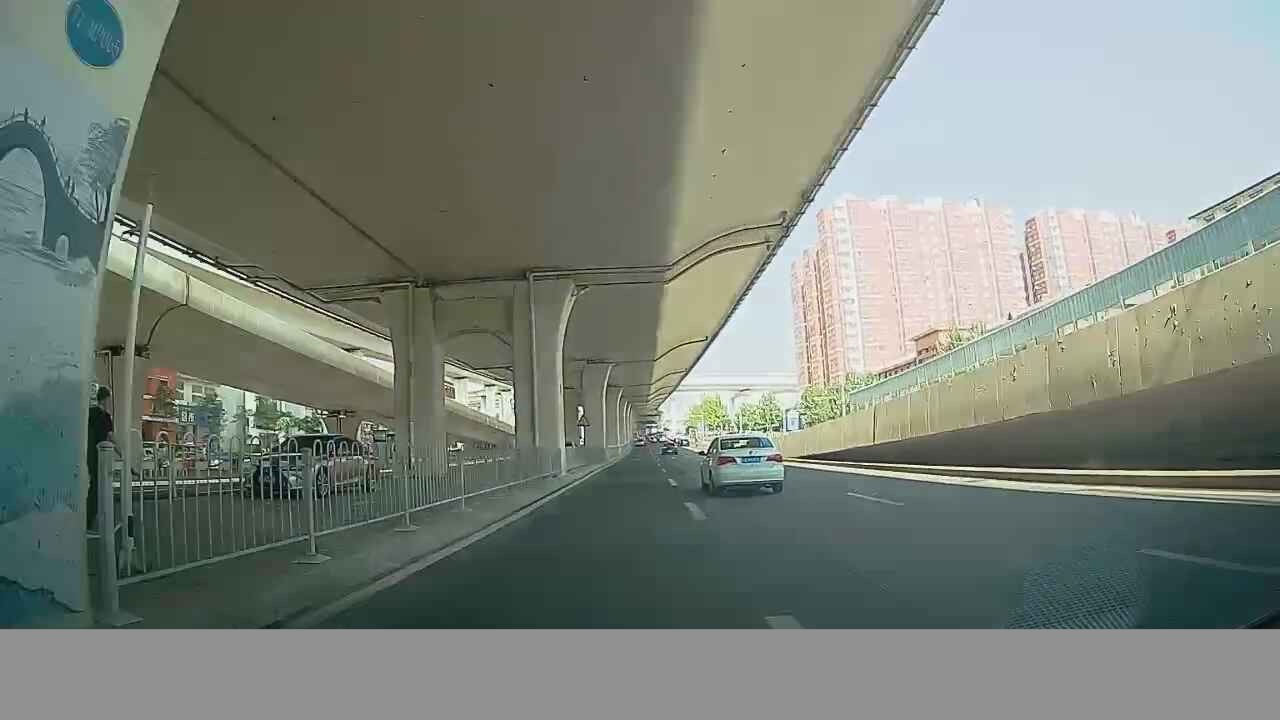}
    \caption{An example image input to MLLM in the FPV road type classification task.}
    \label{fig:A:task6}
\end{figure}

\begin{enhancedpromptbox}
\textbf{Task: Road Type Classification}

\textbf{Image Description:}
This is a personal viewpoint image of a road in China, where vehicles drive on the right side of the road.

\textbf{Question:}
Analyze the image and determine: Is the current vehicle traveling on a main road or a service road?

\textbf{Road Type Definitions:}
\begin{itemize}[leftmargin=1.5em,itemsep=0pt,parsep=0.2em,topsep=0.0em,partopsep=0.0em]
\item \textbf{Main road}: Primary traffic arteries, highways, major streets with higher traffic capacity and speed limits
\item \textbf{Service road}: Secondary roads, access roads, frontage roads, or auxiliary roads that typically serve adjacent properties or provide access to main roads
\end{itemize}

\textbf{YAML Output Requirements:}
\begin{itemize}[leftmargin=1.5em,itemsep=0pt,parsep=0.2em,topsep=0.0em,partopsep=0.0em]
    \item Add ONLY one-line YAML comment explaining your visual analysis and reasoning
    \item Extract \texttt{road\_type} as either \texttt{"main"} or \texttt{"service"}
\end{itemize}

\textbf{Output Format:}
\begin{lstlisting}
```yaml
# Explanation: [Describe what you see that led to your determination]
road_type: <"main" or "service">
```
\end{lstlisting}

Your YAML output:
\end{enhancedpromptbox}

Typical MLLM responses for the image and prompt are as follows:
\begin{enhancedpromptbox}
\textbf{Response from Qwen2.5-VL-32B-instruct:}
\begin{lstlisting}
```yaml
# Explanation: The presence of multiple lanes for vehicular movement, clear lane markings, and an elevated structure (likely part of a highway overpass) suggest this is a primary roadway designed for high-capacity traffic flow.
road_type: main
```
\end{lstlisting}

\textbf{Response from Gemma-3-12B-it:}
\begin{lstlisting}
```yaml
# The presence of multiple lanes, a highway overpass, distinct separation from adjacent buildings, and steady vehicle flow suggest this is a main road.
road_type: main
```
\end{lstlisting}

\end{enhancedpromptbox}

\subsection{Cross-View Lane Counting}

The prompt of the cross-view lane counting task is as follows:

\begin{enhancedpromptbox}
\textbf{Task: Lane Counting for Road Network Analysis}

\textbf{Image Description:}
The first image is a satellite image of a road network in China. A specific road segment has been marked as a red polyline with arrowheads.

The second image is a street-view image of the same road segment. Vehicles drive on the right side of the road. 

\textbf{Question:}
Based on both the satellite image and the street-view image, determine:

How many lanes are available for vehicles traveling in the direction of the arrow?

\textbf{YAML Output Requirements:}
\begin{itemize}[leftmargin=1.5em,itemsep=0pt,parsep=0.2em,topsep=0.0em,partopsep=0.0em]
    \item Add ONLY one-line YAML comment explaining your visual analysis and reasoning
    \item Extract \texttt{num\_lanes} (integer)
\end{itemize}

\textbf{Output Format:}
\begin{lstlisting}
```yaml
# Explanation: [Describe what you see that led to your determination]
num_lanes: <integer>
```
\end{lstlisting}

Your YAML output:
\end{enhancedpromptbox}

Typical MLLM responses for the image and prompt are as follows:
\begin{enhancedpromptbox}
\textbf{Response from Qwen3-VL-32B:}
\begin{lstlisting}
```yaml
# Explanation: The red arrow in the satellite image points south along a road segment that clearly has two lanes in that direction, as seen from the white lane markings and vehicle flow in the street view. The street-view image shows two lanes ahead with vehicles in both, marked by white dashed lines and lane arrows, confirming two lanes for southbound traffic.
num_lanes: 2
```
\end{lstlisting}

\textbf{Response from Qwen2.5-VL-7B-instruct:}
\begin{lstlisting}
```yaml
# Explanation: The red polyline indicates traffic flow from left to right; there appear 3 distinct lane markings visible at this intersection.
num_lanes: 3
```
\end{lstlisting}

\end{enhancedpromptbox}

\subsection{Cross-View Lane Designation Recognition}

An example prompt of the cross lane designation recognition task are as follows:

\begin{figure}[h]
  \centering
    \includegraphics[width=1\linewidth]{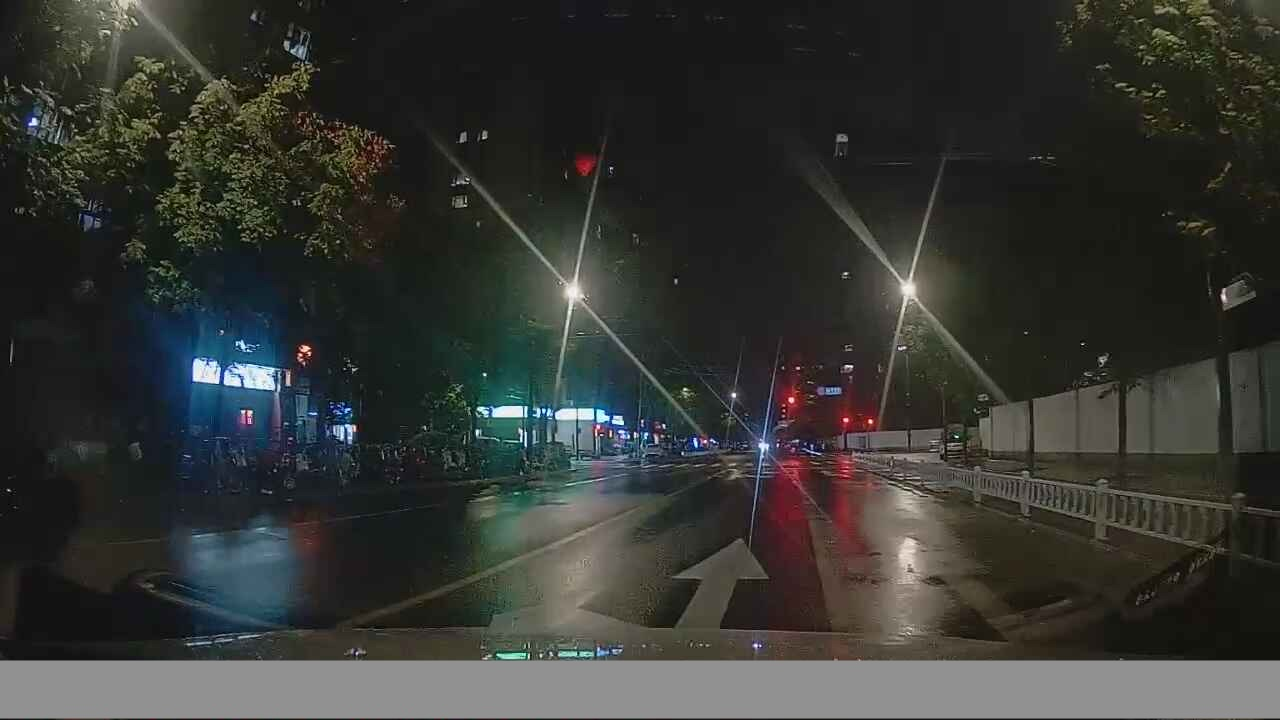}
    \caption{An example image input to MLLM in the FPV lane designation recognition task.}
    \label{fig:A:task5}
\end{figure}

\begin{enhancedpromptbox}
\textbf{Task: Lane Designations Recognition for Road Network Analysis}

\textbf{Image Description:}
The first image is a satellite image of a road network in China. A specific road segment has been marked as a red polyline with arrowheads.

The second image is a personal viewpoint image of a road in China, where vehicles drive on the right side of the road.

\textbf{Data:}
\begin{lstlisting}
```yaml
# The number of lanes
num_lanes: 2
```
\end{lstlisting}

\textbf{Question:}
Analyze the image and determine: What are the lane designations in the direction of the road?

\textbf{Lane Direction Types:}
The available lane direction types include:

(1) U-turn

(2) left-turn

(3) straight

(4) right-turn

(5) variable

\textbf{YAML Output Requirements:}
\begin{itemize}[leftmargin=1.5em,itemsep=0pt,parsep=0.2em,topsep=0.0em,partopsep=0.0em]
    \item Add ONLY one-line YAML comment explaining your visual analysis and reasoning
    \item Return \texttt{lane\_designations} as a list where each item represents the lane direction types for each lane from left to right
    \begin{itemize}
        \item Make sure the number of lane designations is equal to the number of lanes (which is 2)
    \end{itemize}
    \item Each lane can have multiple direction types, so each item in the list should also be a list
    \item Use the English terms: \texttt{"U-turn"}, \texttt{"left-turn"}, \texttt{"straight"}, \texttt{"right-turn"}, \texttt{"variable"}
\end{itemize}

\textbf{Output Format:}
\begin{lstlisting}
```yaml
# Explanation: [Describe what you see that led to your determination]
lane_designations: [["direction1", "direction2"], ["direction3"], ...]
```
\end{lstlisting}

Your YAML output:
\end{enhancedpromptbox}

Typical MLLM responses for the image and prompt are as follows:
\begin{enhancedpromptbox}
\textbf{Response from GLM-4.5V:}
\begin{lstlisting}
```yaml
# Explanation: The satellite image shows a road segment marked by a red polyline indicating direction; given the single lane and typical road usage, the lane is designated for straight movement.
lane_designations: [["straight"]]
```
\end{lstlisting}

\textbf{Response from GPT-5-nano:}
\begin{lstlisting}

# Explanation:  The marked single-lane road shows straight-through traffic in the direction of the arrows; no left/right/U-turn lanes are indicated.
lane_designations: [["straight"]]
\end{lstlisting}

\end{enhancedpromptbox}

%% file: body/A3.result.tex
\section{Additional Experimental Setup Descriptions and Complete Results}\label{sec:A:result}

\subsection{BEV Lane Counting}

The complete experimental results for the BEV lane counting task are presented in Table \ref{tab:task1_results}.
This table reports the performance of each model or baseline method based on Precision, Recall, F1-Score, and Root Mean Square Error (RMSE) metrics.

\begin{table*}[htbp]
\centering
\caption{The complete experimental results of the BEV lane counting task. The best-performing result among the evaluated models is indicated in \textbf{bold}, and the second-best result is indicated with an \underline{underline}.}
\label{tab:task1_results}
\begin{tabular}{lcccc}
\toprule
Model & Precision $\uparrow$ & Recall $\uparrow$ & F1-Score $\uparrow$ & RMSE $\downarrow$ \\
\midrule
LLaMA-3.2-11B-Vision & 0.2655 & 0.252 & 0.2384 & 1.5855 \\
LLaMA-3.2-90B-Vision & 0.2790 & 0.2816 & 0.2947 & 1.3791 \\
Qwen2.5-VL-7B & 0.2628 & 0.2941 & 0.2396 & 1.4164 \\
Qwen2.5-VL-32B & 0.2659 & 0.2910 & 0.2541 & 1.2553 \\
Qwen2.5-VL-72B & 0.2981 & 0.2797 & 0.1999 & 1.3546 \\
Qwen3-VL-8B&0.2481&0.2632&0.2430&1.5068\\
Qwen3-VL-32B&0.3338&0.3292&0.2763&1.2986\\
Qwen3-VL-30B-A3B&0.2587&0.2859&0.2631&1.3187\\
Qwen3-VL-235B-A22B&0.2561&0.2859&0.2565&1.3284\\
Gemma-3-12B & 0.2579 & 0.3137 & 0.2069 & 1.2140 \\
Gemma-3-27B & 0.2923 & 0.3179 & 0.2792 & 1.3342 \\
Gemini-2.5-Flash & 0.2859 & 0.2921 & 0.2452 & 1.3709 \\
Gemini-2.5-Flash-Image & 0.2456 & 0.2797 & 0.2254 & 1.3404 \\
Gemini-2.5-Pro & 0.3089 & 0.3220 & {0.3018} & 1.2524 \\
GLM-4.5V & 0.2901 & \underline{0.3240} & 0.2885 & \underline{1.1830} \\
GPT-5-Nano & 0.2546 & 0.2693 & 0.1966 & 1.5292 \\
GPT-5-Mini & \textbf{0.3782} & \textbf{0.3798} & \textbf{0.3550} & \textbf{1.1623} \\
GPT-5 & \underline{0.3395} & 0.3065 & \underline{0.3048} & 1.3500 \\
\midrule
Random Choice & 0.2403 & 0.2786 & 0.2567 & \textbf{1.4106} \\
Always 2-Lane & 0.0666 & 0.2580 & 0.1058 & 1.5214 \\
\bottomrule
\end{tabular}
\end{table*}

\subsection{BEV Lane Designation Recognition}\label{sec:A:result:task2}

In the lane designation recognition task, we introduce a rule-driven mapping table based on traffic common sense as a baseline:
\begin{itemize}
[leftmargin=1.5em,itemsep=0pt,parsep=0.2em,topsep=0.0em,partopsep=0.0em]
    \item \textbf{1 lane:} The single lane is assigned as \texttt{left-turn}, \texttt{straight}, or \texttt{right-turn}.
    \item \textbf{2 lanes:} The first lane (leftmost) is designated as \texttt{left-turn} or \texttt{straight}, and the second lane (rightmost) as \texttt{straight} or \texttt{right-turn}.
    \item \textbf{3 or more lanes:} The first lane (leftmost) is designated as \texttt{left-turn}, the last lane (rightmost) as \texttt{right-turn}, and all intermediate lanes as \texttt{straight}.
\end{itemize}

Table \ref{tab:task2_results} provides the complete results for the BEV lane designation recognition task. The models and baseline methods are evaluated using Hamming Loss and Accuracy.

\begin{table*}[htbp]
\centering
\caption{The complete experimental results of the BEV lane designation recognition task. The best-performing result among the evaluated models is indicated in \textbf{bold}, and the second-best result is indicated with an \underline{underline}.}
\label{tab:task2_results}
\begin{tabular}{lcc}
\toprule
Model & Hamming Loss $\downarrow$ & Acc. $\uparrow$ \\
\midrule
LLaMA-3.2-11B-Vision & 0.3561 & 0.1337 \\
LLaMA-3.2-90B-Vision & 0.2144 & 0.4572 \\
Qwen2.5-VL-7B & 0.2970 & 0.1107 \\
Qwen2.5-VL-32B & 0.2386 & 0.3609 \\
Qwen2.5-VL-72B & 0.2128 & 0.4630 \\
Qwen3-VL-8B&0.2568&0.3719\\
Qwen3-VL-32B&0.1883&0.5150\\
Qwen3-VL-30B-A3B&0.2722&0.3554\\
Qwen3-VL-235B-A22B&0.1911&0.4957\\
Gemma-3-12B & 0.2563 & 0.3414 \\
Gemma-3-27B & 0.2138 & 0.3994 \\
Gemini-2.5-Flash & 0.2021 & 0.4706 \\
Gemini-2.5-Flash-Image & 0.1857 & 0.4899 \\
Gemini-2.5-Pro & 0.1715 & 0.5235 \\
GLM-4.5V & 0.1751 & 0.5349 \\
GPT-5-Nano & 0.1707 & 0.5404 \\
GPT-5-Mini & \underline{0.1564} & 0.5497 \\
GPT-5 & 0.1592 & \underline{0.5789} \\
\midrule
Rule-based & \textbf{0.1492} & \textbf{0.5872} \\
\bottomrule
\end{tabular}
\end{table*}

\subsection{BEV Road Network Correction}

The comprehensive results for the BEV road network correction task are shown in Table \ref{tab:task3_results}. This task evaluates model performance on junctions by RMSE and road segments by Fréchet Distance (FD) at different normalized distance upper bound thresholds.

\begin{table*}[htbp]
\centering
\small
\caption{The complete experimental results of the BEV road network correction task. The best-performing result among the evaluated models is indicated in \textbf{bold}, and the second-best result is indicated with an \underline{underline}.}
\label{tab:task3_results}
\resizebox{\textwidth}{!}{
\begin{tabular}{lcccccc}
\toprule
\multirow{2}{*}{Model} & \multicolumn{3}{c}{Junction} & \multicolumn{3}{c}{Road Segment} \\
\cmidrule(lr){2-7}
 & RMSE@10\% $\downarrow$ & RMSE@20\% $\downarrow$ & RMSE@50\% $\downarrow$ & FD@10\% $\downarrow$ & FD@20\% $\downarrow$ & FD@50\% $\downarrow$ \\
\midrule
LLaMA-3.2-11B-Vision & 0.0938 & 0.1725 & 0.3392 & 0.0977 & 0.1876 & 0.3772 \\
LLaMA-3.2-90B-Vision & 0.0842 & 0.1483 & 0.2730 & 0.0955 & 0.1749 & 0.3062 \\
Qwen2.5-VL-7B & 0.0924 & 0.1697 & 0.3455 & 0.0978 & 0.1862 & 0.3988 \\
Qwen2.5-VL-32B & 0.0874 & 0.1553 & 0.3147 & 0.0954 & 0.1722 & 0.3137 \\
Qwen2.5-VL-72B & \underline{0.0815} & \underline{0.1387} & 0.2668 & 0.0921 & 0.1603 & 0.2677 \\
Qwen3-VL-8B & 0.0861 & 0.1513 & 0.3004 & 0.0942 & 0.1680 & 0.2938 \\
Qwen3-VL-32B & 0.0851 & 0.1450 & 0.2847 & 0.0932 & 0.1611 & 0.2567 \\
Qwen3-VL-30B-A3B & 0.0843 & 0.1476 & 0.2941 & 0.0948 & 0.1732 & 0.3020 \\
Qwen3-VL-235B-A22B & \underline{0.0815} & 0.1389 & 0.2657 & 0.0908 & 0.1574 & 0.2495 \\
Gemma-3-12B & 0.0872 & 0.1560 & 0.3106 & 0.0936 & 0.1677 & 0.2894 \\
Gemma-3-27B & 0.0860 & 0.1457 & 0.2785 & 0.0931 & 0.1553 & \textbf{0.2319} \\
Gemini-2.5-Flash & 0.0845 & 0.1481 & 0.3174 & 0.0962 & 0.1725 & 0.2781 \\
Gemini-2.5-Flash-Image & 0.0848 & 0.1469 & 0.3050 & 0.0962 & 0.1686 & 0.2618 \\
Gemini-2.5-Pro & 0.0834 & 0.1400 & 0.2909 & 0.0957 & 0.1725 & 0.2894 \\
GLM-4.5V & 0.0910 & 0.1708 & 0.3986 & \textbf{0.0892} & \underline{0.1525} & 0.2676 \\
GPT-5-Nano & \textbf{0.0806} & \textbf{0.1316} & \textbf{0.2570} & 0.0918 & 0.1552 & 0.2582 \\
GPT-5-Mini & 0.0830 & 0.1407 & 0.2920 & \underline{0.0905} & \textbf{0.1479} & \underline{0.2346} \\
GPT-5 & 0.0853 & 0.1430 & 0.2907 & 0.0950 & 0.1635 & 0.2608 \\
\midrule 
Do Nothing & 0.0850 & 0.1555 & \underline{0.2609} & 0.0996 & 0.1944 & 0.3793 \\
\bottomrule
\end{tabular}
}
\end{table*}
\subsection{FPV Lane Counting}

The results for the FPV lane counting task are summarized in Table \ref{tab:task4_results}. The table includes Precision, Recall, F1-Score, and RMSE for each model and baseline method.

\begin{table*}[htbp]
\centering
\caption{The complete experimental results of the FPV lane counting task. The best-performing result among the evaluated models is indicated in \textbf{bold}, and the second-best result is indicated with an \underline{underline}.}
\label{tab:task4_results}
\begin{tabular}{lcccc}
\toprule
Model & Precision $\uparrow$ & Recall $\uparrow$ & F1-Score $\uparrow$ & RMSE $\downarrow$ \\
\midrule
LLaMA-3.2-11B-Vision & 0.2494 & 0.2696 & 0.2439 & 1.4978 \\
LLaMA-3.2-90B-Vision & 0.4156 & 0.3652 & 0.3094 & 1.1647 \\
Qwen2.5-VL-7B & 0.2717 & 0.2522 & 0.2030 & 1.5548 \\
Qwen2.5-VL-32B & 0.3940 & 0.3652 & 0.3072 & 1.0994 \\
Qwen2.5-VL-72B & \textbf{0.5883} & 0.3739 & 0.3176 & 1.2649 \\
Qwen3-VL-8B & 0.2203 & 0.3652 & 0.2664 & 1.3221 \\
Qwen3-VL-32B & 0.5286 & 0.4087 & 0.3400 & 1.1151 \\
Qwen3-VL-30B-A3B & 0.4442 & 0.3478 & 0.2865 & 1.2087 \\
Qwen3-VL-235B-A22B & 0.4991 & 0.4348 & 0.3690 & 1.0342 \\
Gemma-3-12B & 0.3940 & 0.2783 & 0.2000 & 1.2123 \\
Gemma-3-27B & 0.3422 & 0.2957 & 0.2498 & 1.2854 \\
Gemini-2.5-Flash & 0.4238 & 0.3652 & 0.3270 & 1.2158 \\
Gemini-2.5-Flash-Image & 0.2844 & 0.2870 & 0.2251 & 1.2683 \\
Gemini-2.5-Pro & \underline{0.5796} & \textbf{0.5304} & \textbf{0.5224} & \textbf{0.8022} \\
GLM-4.5V & 0.4775 & 0.3565 & 0.3219 & 1.0714 \\
GPT-5-Nano & 0.1712 & 0.3043 & 0.2112 & 1.3449 \\
GPT-5-Mini & 0.4168 & 0.4087 & 0.3619 & 1.0087 \\
GPT-5 & 0.5317 & \underline{0.5217} & \underline{0.4983} & \underline{0.9231} \\
\midrule
Random Choice & 0.2661 & 0.3217 & 0.2842 & 1.4295 \\
Always 2-Lane & 0.0473 & 0.2174 & 0.0776 & 1.6965 \\
\bottomrule
\end{tabular}
\end{table*}
\subsection{FPV Lane Designation Recognition}

Table \ref{tab:task5_results} displays the complete experimental outcomes for the FPV lane designation recognition task, measured by Hamming Loss and Accuracy.

\begin{table}[htbp]
\centering
\caption{The complete experimental results of the FPV lane designation recognition task. The best-performing result among the evaluated models is indicated in \textbf{bold}, and the second-best result is indicated with an \underline{underline}.}
\label{tab:task5_results}
\begin{tabular}{lcc}
\toprule
Model & Hamming Loss $\downarrow$ & Acc. $\uparrow$ \\
\midrule
LLaMA-3.2-11B-Vision & 0.3237 & 0.1793 \\
LLaMA-3.2-90B-Vision & 0.1657 & 0.5137 \\
Qwen2.5-VL-7B & 0.2781 & 0.1581 \\
Qwen2.5-VL-32B & 0.2363 & 0.3100 \\
Qwen2.5-VL-72B & 0.1877 & 0.4712 \\
Qwen3-VL-8B & 0.2082 & 0.4407 \\
Qwen3-VL-32B & 0.1444 & 0.5866 \\
Qwen3-VL-30B-A3B & 0.2242 & 0.4286 \\
Qwen3-VL-235B-A22B & 0.1588 & 0.5410 \\
Gemma-3-12B & 0.2629 & 0.3435 \\
Gemma-3-27B & 0.1945 & 0.4043 \\
Gemini-2.5-Flash & 0.1755 & 0.4802 \\
Gemini-2.5-Flash-Image & 0.1733 & 0.4711 \\
Gemini-2.5-Pro & 0.1603 & 0.5471 \\
GLM-4.5V & 0.1451 & 0.5714 \\
GPT-5-Nano & 0.1512 & 0.5502 \\
GPT-5-Mini & 0.1368 & 0.5836 \\
GPT-5 & \textbf{0.1216} & \textbf{0.6261} \\
\midrule
Rule-based & \underline{0.1269} & \underline{0.6049} \\
\bottomrule
\end{tabular}
\end{table}
\subsection{FPV Road Type Classification}

The performance of all models on the FPV road type classification task is detailed in Table \ref{tab:task6_result}, with results reported in Accuracy.

\begin{table}[htbp]
\centering
\caption{The complete experimental results of the FPV road type classification task. The best-performing result among the evaluated models is indicated in \textbf{bold}, and the second-best result is indicated with an \underline{underline}.}
\label{tab:task6_result}
\begin{tabular}{lc}
\toprule
Model & Acc. $\uparrow$ \\
\midrule
LLaMA-3.2-11B-Vision & 0.5273 \\
LLaMA-3.2-90B-Vision & 0.6545 \\
Qwen2.5-VL-7B & 0.5152 \\
Qwen2.5-VL-32B & 0.5424 \\
Qwen2.5-VL-72B & 0.5848 \\
Qwen3-VL-8B &0.5394\\
Qwen3-VL-32B &0.5333\\
Qwen3-VL-30B-A3B &0.5333\\
Qwen3-VL-235B-A22B &0.5303\\
Gemma-3-12B & 0.5485 \\
Gemma-3-27B & 0.5121 \\
Gemini-2.5-Flash & 0.5364 \\
Gemini-2.5-Flash-Image & 0.5545 \\
Gemini-2.5-Pro & \textbf{0.7485} \\
GLM-4.5V & 0.5697 \\
GPT-5-Nano & 0.6091 \\
GPT-5-Mini & 0.5879 \\
GPT-5 & \underline{0.6848} \\
\midrule
Random Choice & 0.4576 \\
\bottomrule
\end{tabular}
\end{table}

\subsection{Cross-View Lane Counting}

The results for the cross-view lane counting task are summarized in Table \ref{tab:task7_results}. The table includes Precision, Recall, F1-Score, and RMSE for each model and baseline method.

\begin{table*}[htbp]
\centering
\caption{The complete experimental results of the cross-view lane counting task. The best-performing result among the evaluated models is indicated in \textbf{bold}, and the second-best result is indicated with an \underline{underline}.}
\label{tab:task7_results}
\begin{tabular}{lcccc}
\toprule
Model & Precision $\uparrow$ & Recall $\uparrow$ & F1-Score $\uparrow$ & RMSE $\downarrow$ \\
\midrule
LLaMA-3.2-11B-Vision & - & - & - & - \\
LLaMA-3.2-90B-Vision & - & - & - & - \\
Qwen2.5-VL-7B & 0.2177 & 0.2560 & 0.2325 & 1.3176 \\
Qwen2.5-VL-32B & 0.2388 & 0.2823 & 0.2265 & 1.2476 \\
Qwen2.5-VL-72B & \textbf{0.4974} & 0.3206 & 0.2388 & 1.3842 \\
Qwen3-VL-8B & 0.2279 & 0.3588 & 0.2662 & 1.1819 \\
Qwen3-VL-32B & 0.4213 & 0.3893 & 0.2976 & 1.1851 \\
Qwen3-VL-30B-A3B & 0.2900 & 0.3130 & 0.2787 & 1.3135 \\
Qwen3-VL-235B-A22B & 0.3959 & 0.3511 & 0.2981 & 1.3164 \\
Gemma-3-12B & 0.2155 & 0.2901 & 0.1961 & 1.2011 \\
Gemma-3-27B & 0.2245 & 0.3282 & 0.2483 & 1.1392 \\
Gemini-2.5-Flash & 0.3092 & 0.3053 & 0.2338 & 1.1819 \\
Gemini-2.5-Flash-Image & - & - & - & - \\
Gemini-2.5-Pro & 0.3876 & 0.3817 & \underline{0.3649} & 1.0947 \\
GLM-4.5V & 0.3788 & 0.3893 & 0.3289 & 1.3394 \\
GPT-5-Nano & 0.3440 & 0.2901 & 0.2239 & 1.3842 \\
GPT-5-Mini & \underline{0.4827} & \underline{0.3969} & 0.3437 & \underline{1.0114} \\
GPT-5 & 0.4412 & \textbf{0.4122} & \textbf{0.3996} & \textbf{0.8781} \\
\midrule
Random Choice & 0.2370 & 0.2748 & 0.2516 & 1.3952 \\
Always 2-Lane & 0.0490 & 0.2214 & 0.0802 & 1.6228 \\
\bottomrule
\end{tabular}
\end{table*}
\subsection{Cross-View Lane Designation Recognition}

Table \ref{tab:task8_results} displays the complete experimental outcomes for the cross-view lane designation recognition task, measured by Hamming Loss and Accuracy.

\begin{table}[htbp]
\centering
\caption{The complete experimental results of the cross-view lane designation recognition task. The best-performing result among the evaluated models is indicated in \textbf{bold}, and the second-best result is indicated with an \underline{underline}.}
\label{tab:task8_results}
\begin{tabular}{lcc}
\toprule
Model & Hamming Loss $\downarrow$ & Acc. $\uparrow$ \\
\midrule
LLaMA-3.2-11B-Vision & - & - \\
LLaMA-3.2-90B-Vision & - & - \\
Qwen2.5-VL-7B & 0.2734 & 0.2027 \\
Qwen2.5-VL-32B & 0.2060 & 0.4076 \\
Qwen2.5-VL-72B & 0.1832 & 0.4566 \\
Qwen3-VL-8B & 0.2171 & 0.4321 \\
Qwen3-VL-32B & 0.1754 & 0.5145 \\
Qwen3-VL-30B-A3B & 0.2233 & 0.4076 \\
Qwen3-VL-235B-A22B & 0.1782 & 0.4788 \\
Gemma-3-12B & 0.2606 & 0.3474 \\
Gemma-3-27B & 0.2238 & 0.3808 \\
Gemini-2.5-Flash & 0.2823 & 0.0935 \\
Gemini-2.5-Flash-Image & - & - \\
Gemini-2.5-Pro & 0.2823 & 0.0935 \\
GLM-4.5V & 0.1422 & 0.5569 \\
GPT-5-Nano & 0.1531 & 0.5256 \\
GPT-5-Mini & \textbf{0.1214} & \underline{0.5880} \\
GPT-5 & \underline{0.1236} & \textbf{0.5969} \\
\midrule
Rule-based & 0.1303 & \underline{0.5880} \\
\bottomrule
\end{tabular}
\end{table}

\subsection{Uncertainty-aware Evaluation on Small Splits}
\label{app:small_split_uncertainty}

Some evaluation splits in RoadBench, especially FPV lane designation and cross-view tasks, contain relatively fewer test cases due to the difficulty of collecting high-quality geographically paired BEV-FPV samples. To assess the statistical reliability of the reported results, we conduct three repeated evaluations for representative frontier models on these small splits and report the mean and standard deviation of each metric. As shown in Table~\ref{tab:compact_uncertainty_comparison}, the overall conclusion remains stable: current MLLMs still struggle with fine-grained FPV and cross-view lane reasoning. Meanwhile, the non-negligible variance on some metrics indicates that small pairwise differences between frontier models should not be over-interpreted as definitive rankings. Therefore, in the main paper, we emphasize consistent trends across models and tasks rather than marginal performance gaps on small splits.

\begin{table}[t]
\centering
\scriptsize
\setlength{\tabcolsep}{2pt}
\renewcommand{\arraystretch}{0.84}
\begin{tabular}{@{}lcccccc@{}}
\toprule
& \multicolumn{2}{c}{FPV-LD}
& \multicolumn{2}{c}{CV-LD}
& \multicolumn{2}{c}{CV-LC} \\
\cmidrule(lr){2-3}\cmidrule(lr){4-5}\cmidrule(lr){6-7}
Model & HL$\downarrow$ & Acc.$\uparrow$ & HL$\downarrow$ & Acc.$\uparrow$ & F1$\uparrow$ & RMSE$\downarrow$ \\
\midrule
Gemini-Pro
& .14$\pm$.01 & .58$\pm$.03
& .20$\pm$.08 & .39$\pm$.26
& .37$\pm$.01 & 1.02$\pm$.08 \\
GPT-5-Mini
& .12$\pm$.02 & .62$\pm$.04
& .12$\pm$.01 & .59$\pm$.02
& .34$\pm$.00 & 1.13$\pm$.23 \\
\bottomrule
\end{tabular}
\caption{Uncertainty-aware results on small evaluation splits. We report the mean and standard deviation over three repeated evaluations.}
\label{tab:compact_uncertainty_comparison}
\vspace{-10pt}
\end{table}

%% file: body/A4.further.tex
\section{Results of Further Analysis}\label{sec:A:further}

\subsection{The Impact of Reference Line Prompting Methods in the BEV Tasks}

The experimental results of the impact of reference line prompting methods in the BEV tasks are shown in Table~\ref{tab:glm_prompt_analysis} and Table~\ref{tab:gpt_prompt_analysis}.

\begin{table*}[htbp]
\small
\centering
\caption{GLM-4.5V performance comparison with different prompt strategies. The best-performing result among the evaluated models is indicated in \textbf{bold}, and the second-best result is indicated with an \underline{underline}.}
\label{tab:glm_prompt_analysis}
\resizebox{\textwidth}{!}{
\begin{tabular}{lcccc}
\toprule
\multirow{2}{*}{Method} & \multicolumn{2}{c}{BEV Lane Counting} & \multicolumn{2}{c}{BEV Lane Designation Recognition} \\
\cmidrule(lr){2-5}
 & F1-Score $\uparrow$ & RMSE $\downarrow$ & Hamming Loss $\downarrow$ & Acc. $\uparrow$ \\
\midrule
Text-only & \textbf{0.3017}  & 1.3802 & 0.1906 & 0.4854 \\
Visual-only (Color) & \textbf{0.3017} & 1.1998  & 0.1811  & 0.5167  \\
Visual-only (Arrow) & 0.2920  & \textbf{1.1417} & \underline{0.1764 } & \underline{0.5301}  \\
Both (Color) & \underline{0.3011} & 1.2966  & 0.1794 & {0.5170 } \\
Both (Arrow) & {0.2885 } & \underline{1.1830 } & \textbf{0.1751 } & \textbf{0.5349 } \\
\bottomrule
\end{tabular}
}
\end{table*}

\begin{table*}[htbp]
\small
\centering
\caption{GPT-5-Mini performance comparison with different prompt strategies. The best-performing result among the evaluated models is indicated in \textbf{bold}, and the second-best result is indicated with an \underline{underline}.}
\label{tab:gpt_prompt_analysis}
\resizebox{\textwidth}{!}{
\begin{tabular}{lcccc}
\toprule
\multirow{2}{*}{Method} & \multicolumn{2}{c}{BEV Lane Counting} & \multicolumn{2}{c}{BEV Lane Designation Recognition} \\
\cmidrule(lr){2-5}
 & F1-Score $\uparrow$ & RMSE $\downarrow$ & Hamming Loss $\downarrow$ & Acc. $\uparrow$ \\
\midrule
Text-only & 0.3145 & 1.2741 & 0.1535 & 0.5514 \\
Visual-only (Color) & {0.3378} & 1.1921 & {0.1526} & 0.5497 \\
Visual-only (Arrow) & 0.3086 & 1.1916 & \underline{0.1513} & \textbf{0.5634} \\
Both (Color) & \underline{0.3379} & \textbf{1.1371} & \textbf{0.1503} & \underline{0.5593} \\
Both (Arrow) & \textbf{0.3550} & \underline{1.1623} & 0.1564 & 0.5497 \\
\bottomrule
\end{tabular}
}
\end{table*}

\subsection{The Impact of Scene Environment Conditions in the FPV Tasks}

The experimental results of the impact of scene environment conditions in the FPV tasks are shown in Table~\ref{tab:glm4v_analysis}, Table~\ref{tab:gpt_analysis}, and Table~\ref{tab:gemini_analysis}.

\begin{table*}[htbp]
\small
\centering
\caption{GLM-4.5V performance analysis across different environmental conditions.}
\label{tab:glm4v_analysis}
\resizebox{\textwidth}{!}{
\begin{tabular}{lcccc}
\toprule
\multirow{2}{*}{Condition} & \multicolumn{2}{c}{FPV Lane Counting} & \multicolumn{2}{c}{FPV Lane Designation Recognition} \\
\cmidrule(lr){2-5}
 & F1-Score $\uparrow$ & RMSE $\downarrow$ & Hamming Loss $\downarrow$ & Accuracy $\uparrow$ \\
\midrule
Adverse Lighting Conditions& 0.2076&0.9562&0.1360&0.5294 \\
Obscured Road Markings & 0.3056&1.291& 0.1161&0.5357\\
\textbf{All}& 0.3219& 1.0714&0.1451& 0.5714 \\
\bottomrule
\end{tabular}
}
\end{table*}

\begin{table*}[htbp]
\small
\centering
\caption{GPT-5 performance analysis across different environmental conditions.}
\label{tab:gpt_analysis}
\resizebox{\textwidth}{!}{
\begin{tabular}{lcccc}
\toprule
\multirow{2}{*}{Condition} & \multicolumn{2}{c}{FPV Lane Counting} & \multicolumn{2}{c}{FPV Lane Designation Recognition} \\
\cmidrule(lr){2-5}
 & F1-Score $\uparrow$ & RMSE $\downarrow$ & Hamming Loss $\downarrow$ & Accuracy $\uparrow$ \\
\midrule
Adverse Lighting Conditions & 0.5744&0.7559 & 0.0882& 0.7059 \\
Obscured Road Markings &0.3056 &1.291 & 0.1250&0.6071 \\
\textbf{All}& 0.4983 &0.9231 &0.1216 &0.6261 \\
\bottomrule
\end{tabular}
}
\end{table*}

\begin{table*}[htbp]
\small
\centering
\caption{Gemini-2.5-Pro performance analysis across different environmental conditions.}
\label{tab:gemini_analysis}
\resizebox{\textwidth}{!}{
\begin{tabular}{lcccc}
\toprule
\multirow{2}{*}{Condition} & \multicolumn{2}{c}{FPV Lane Counting} & \multicolumn{2}{c}{FPV Lane Designation Recognition} \\
\cmidrule(lr){2-5}
 & F1-Score $\uparrow$ & RMSE $\downarrow$ & Hamming Loss $\downarrow$ & Accuracy $\uparrow$ \\
\midrule
Adverse Lighting Conditions & 0.5471&0.8106 &0.1324&0.5882 \\
Obscured Road Markings &0.7778&0.4082 &0.1161&0.5714 \\
\textbf{All}& 0.5224 & 0.8022&0.1603 &0.5471 \\
\bottomrule
\end{tabular}
}
\end{table*}

%% file: body/A5.case.tex
\newpage

\newpage

\section{Case Analysis}\label{sec:A:case}

This section presents a detailed analysis of failure cases across different tasks in RoadBench, highlighting common failure patterns and limitations of current MLLMs in fine-grained spatial understanding and reasoning under urban road scenarios.
In the following analyzes, we selected the evaluation results of the MLLMs that ranked highly in the benchmarks corresponding to the tasks to be analyzed.
In the text description of each failure case, the text in the \textbf{Analysis} paragraph is a brief error description provided after manual checking, not a model output.

\subsection{BEV Lane Counting}

This subsection presents failure cases generated by GPT-5-Mini in BEV lane counting tasks.

\begin{figure}[h]
  \centering
  \begin{subfigure}[b]{0.32\linewidth}
    \includegraphics[width=\linewidth]{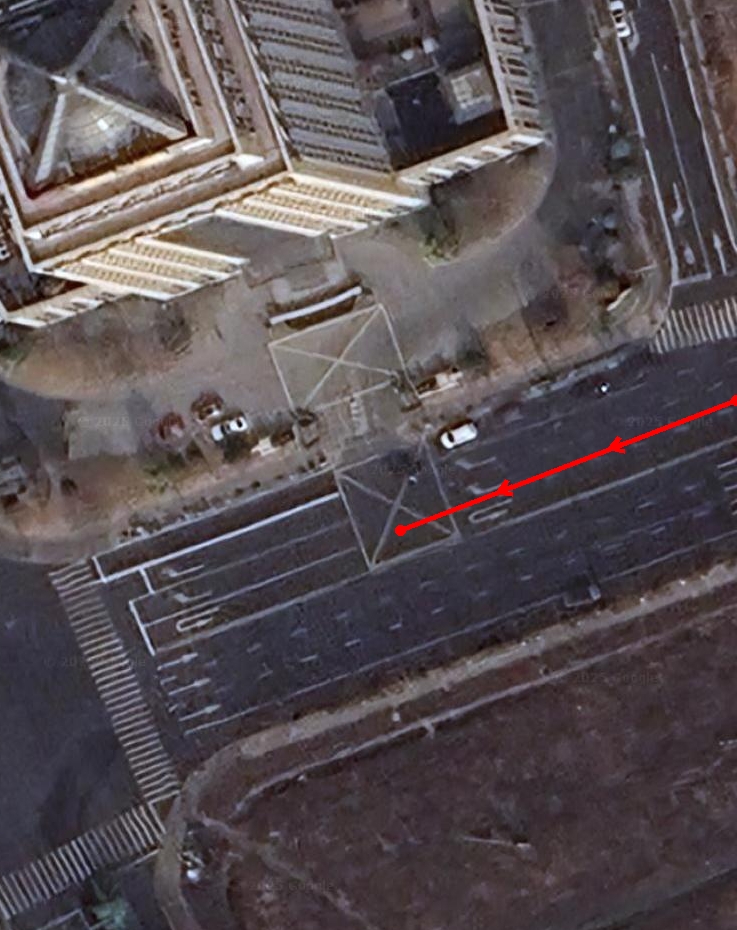}
    \caption{Case 1}
  \end{subfigure}
  \hfill
  \begin{subfigure}[b]{0.32\linewidth}
    \includegraphics[width=\linewidth]{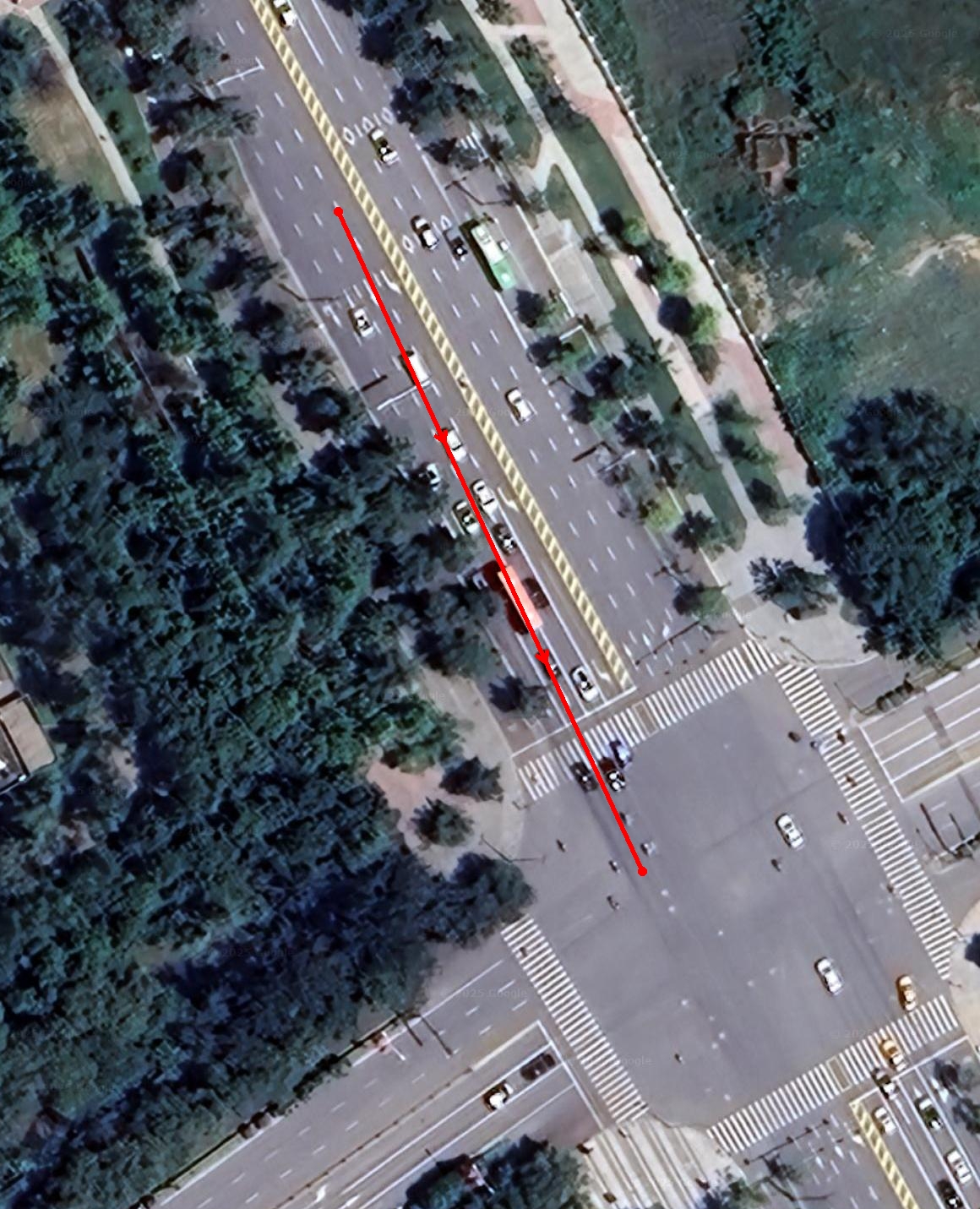}
    \caption{Case 2}
  \end{subfigure}
  \hfill
  \begin{subfigure}[b]{0.32\linewidth}
    \includegraphics[width=\linewidth]{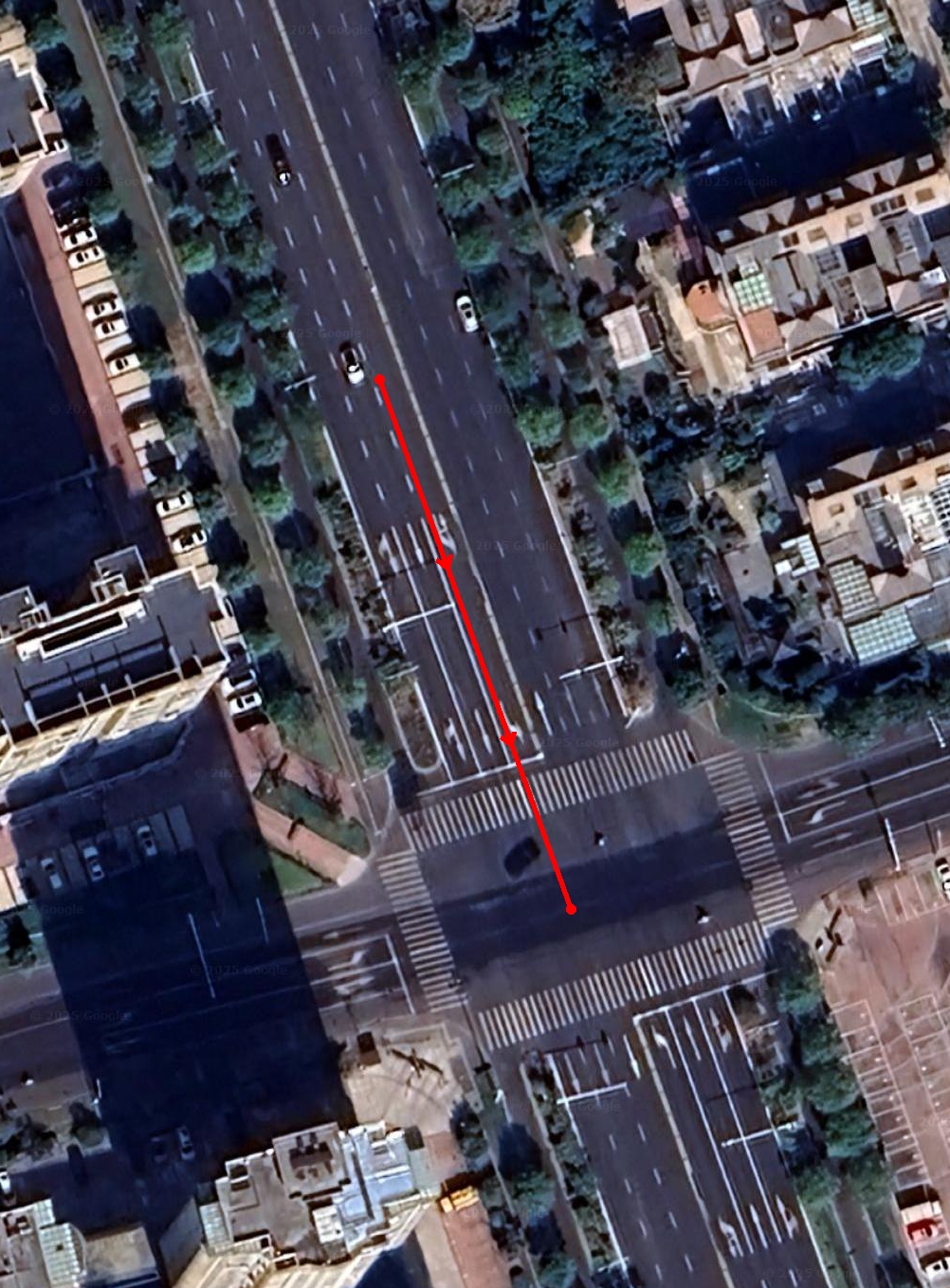}
    \caption{Case 3}
  \end{subfigure}
  \caption{Example images of BEV lane counting failure cases.}
  \label{fig:badcases:task1}
\end{figure}

\begin{enhancedpromptbox}

\textbf{Case 1:}

\textbf{MLLM Response:}

\begin{lstlisting}
```yaml
# Explanation: I counted the distinct through-lane markings and dashed separators in the marked carriageway - there are four parallel through lanes for traffic moving in the arrow direction
num_lanes: 4
```
\end{lstlisting}

\textbf{Ground Truth:}

\begin{lstlisting}
```yaml
num_lanes: 3
```
\end{lstlisting}

\textbf{Analysis:} The model failed to infer that the rightmost lane is a non-motorized vehicle lane based on the presence of ground markings, leading to an overcount of lanes.

\end{enhancedpromptbox}

\begin{enhancedpromptbox}

\textbf{Case 2:}

\textbf{MLLM Response:}

\begin{lstlisting}
```yaml
# Explanation: The marked carriageway (right-side traffic flow) shows three distinct lane markings/queues (three rows of vehicles and two dashed separators) indicating three lanes in the arrow direction
num_lanes: 3
```
\end{lstlisting}

\textbf{Ground Truth:}

\begin{lstlisting}
```yaml
num_lanes: 4
```
\end{lstlisting}

\textbf{Analysis:} The model failed to reason that the leftmost lane, despite having no vehicles, is still a motor vehicle lane, resulting in an undercount.

\end{enhancedpromptbox}

\begin{enhancedpromptbox}

\textbf{Case 3:}

\textbf{MLLM Response:}

\begin{lstlisting}
```yaml
# Explanation: I inspected the carriageway on the side corresponding to the arrow (right-hand driving). I can see four distinct parallel traffic lanes heading in the arrow direction (multiple through lanes plus a rightmost turn/through lane) based on the painted lane markings and turn bays.
num_lanes: 4
```
\end{lstlisting}

\textbf{Ground Truth:}

\begin{lstlisting}
```yaml
num_lanes: 3
```
\end{lstlisting}

\textbf{Analysis:} The model demonstrates poor understanding of bidirectional roads, incorrectly counting lanes from the opposite direction's arrow markings as part of the current direction's lane count.

\end{enhancedpromptbox}

\textbf{Summary:} In this task, MLLMs show their shortcomings in understanding and reasoning about the identification of fine-grained spatial elements and the associations between elements. Case 3 exemplifies the failure of MLLMs to recognize arrows in the opposite direction. Case 1 and Case 2, on the other hand, show that MLLMs are unable to effectively integrate the complex relationship between vehicles and road markings.

\subsection{BEV Lane Designation Recognition}

This subsection presents failure cases generated by GPT-5 in BEV lane designation recognition tasks.

\begin{figure}[h]
  \centering
  \begin{subfigure}[b]{0.48\linewidth}
    \includegraphics[width=\linewidth]{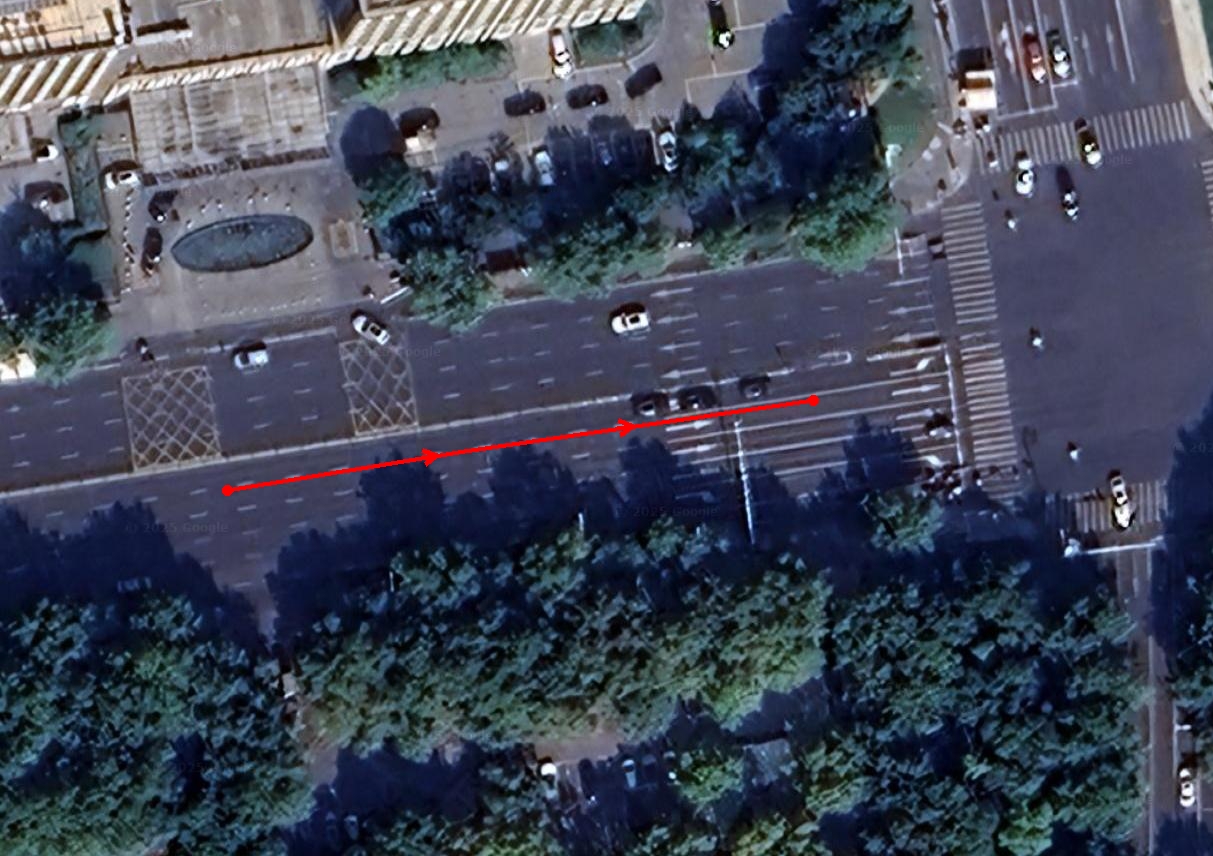}
    \caption{Case 1}
  \end{subfigure}
  \hfill
  \begin{subfigure}[b]{0.48\linewidth}
    \includegraphics[width=\linewidth]{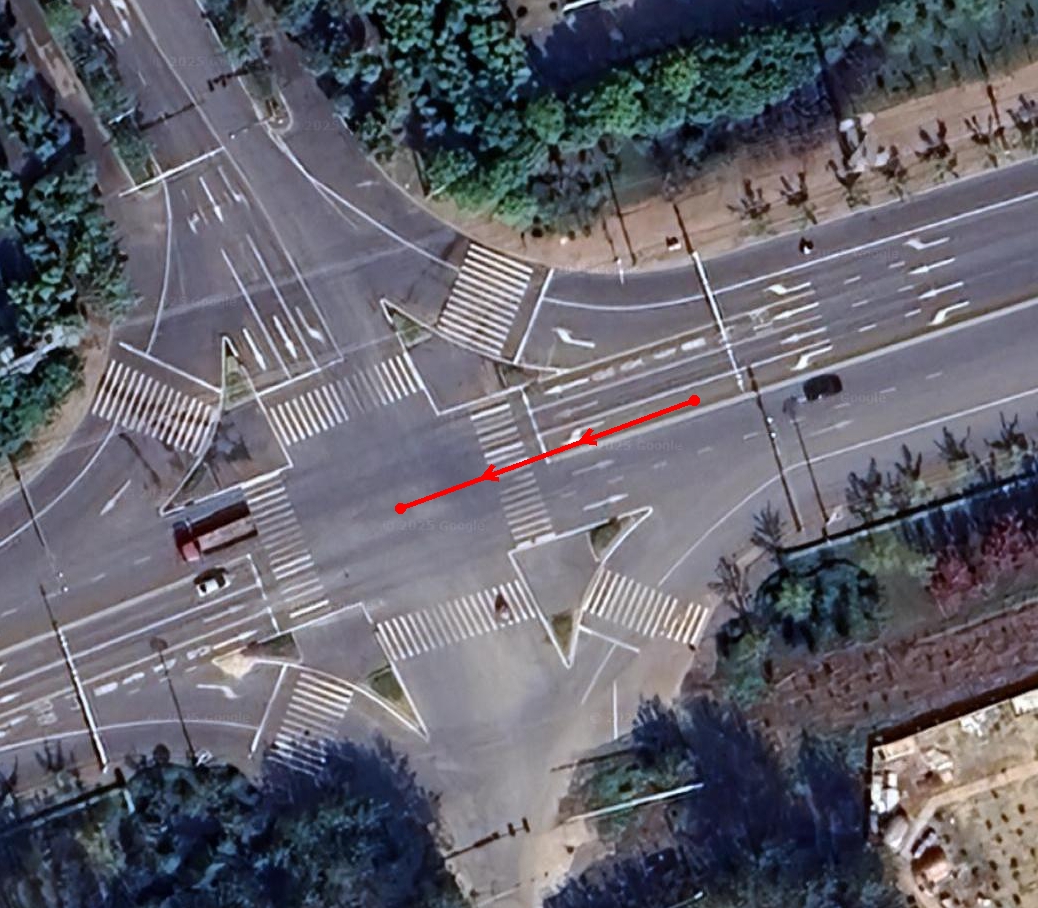}
    \caption{Case 2}
  \end{subfigure}
  \caption{Example images of BEV lane designation failure cases.}
  \label{fig:badcases:task2}
\end{figure}

\begin{enhancedpromptbox}

\textbf{Case 1:}

\textbf{MLLM Response:}

\begin{lstlisting}
```yaml
# Explanation: Near the intersection, lane arrows are visible-leftmost shows a left-turn, the next shows both left-turn and straight, the middle two show straight-only arrows, and the far-right lane shows a right-turn arrow.
lane_designations: [["left-turn"], ["left-turn", "straight"], ["straight"], ["straight"], ["right-turn"]]
```
\end{lstlisting}

\textbf{Ground Truth:}

\begin{lstlisting}
```yaml
lane_designations: [["U-turn", "left-turn"], ["straight"], ["straight"], ["straight"], ["right-turn"]]
```
\end{lstlisting}

\textbf{Analysis:} The model failed to identify a U-turn lane that is located before the junction, demonstrating insufficient understanding of transportation system markings and road design patterns.

\end{enhancedpromptbox}

\begin{enhancedpromptbox}

\textbf{Case 2:}

\textbf{MLLM Response:}

\begin{lstlisting}
```yaml
# Explanation: On the east approach (right side) the three lanes next to a separate channelized right-turn slip lane show straight arrows, with left-turn lanes segregated by islands-so the through carriageway is straight-only.
lane_designations: [["straight"], ["straight"], ["straight"]]
```
\end{lstlisting}

\textbf{Ground Truth:}

\begin{lstlisting}
```yaml
lane_designations: [["left-turn"], ["straight"], ["straight"]]
```
\end{lstlisting}

\textbf{Analysis:} The model failed to understand that this is a bidirectional road and incorrectly searched for left-turn lanes in the lower portion of the image instead of recognizing the left-turn lane in the current direction.

\end{enhancedpromptbox}

\textbf{Summary:} In this task, MLLMs similarly demonstrate a lack of fine-grained road marking recognition (Case 1) and a lack of understanding of the road marking system (Case 2).

\subsection{BEV Road Network Correction}

This subsection presents failure cases generated by Gemma-3-27B in BEV road network correction tasks.
In the images for this task, the red lines represent the inputs to the MLLMs, the blue lines represent the outputs of the MLLMs, and the green lines represent the truth values labeled in the dataset.

\begin{figure}[h]
  \centering
  \begin{subfigure}[b]{0.32\linewidth}
    \includegraphics[width=\linewidth]{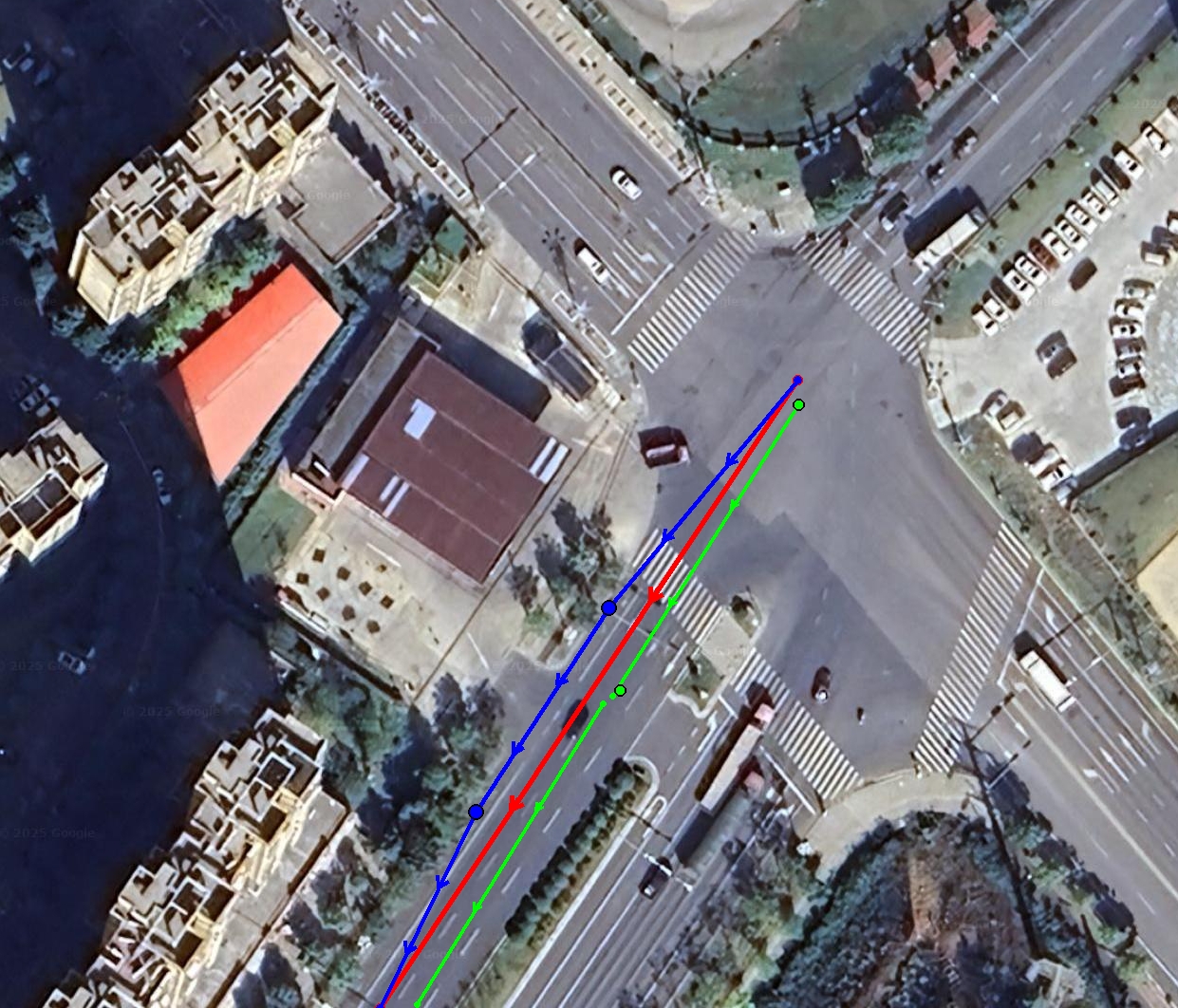}
    \caption{Case 1}
  \end{subfigure}
  \hfill
  \begin{subfigure}[b]{0.32\linewidth}
    \includegraphics[width=\linewidth]{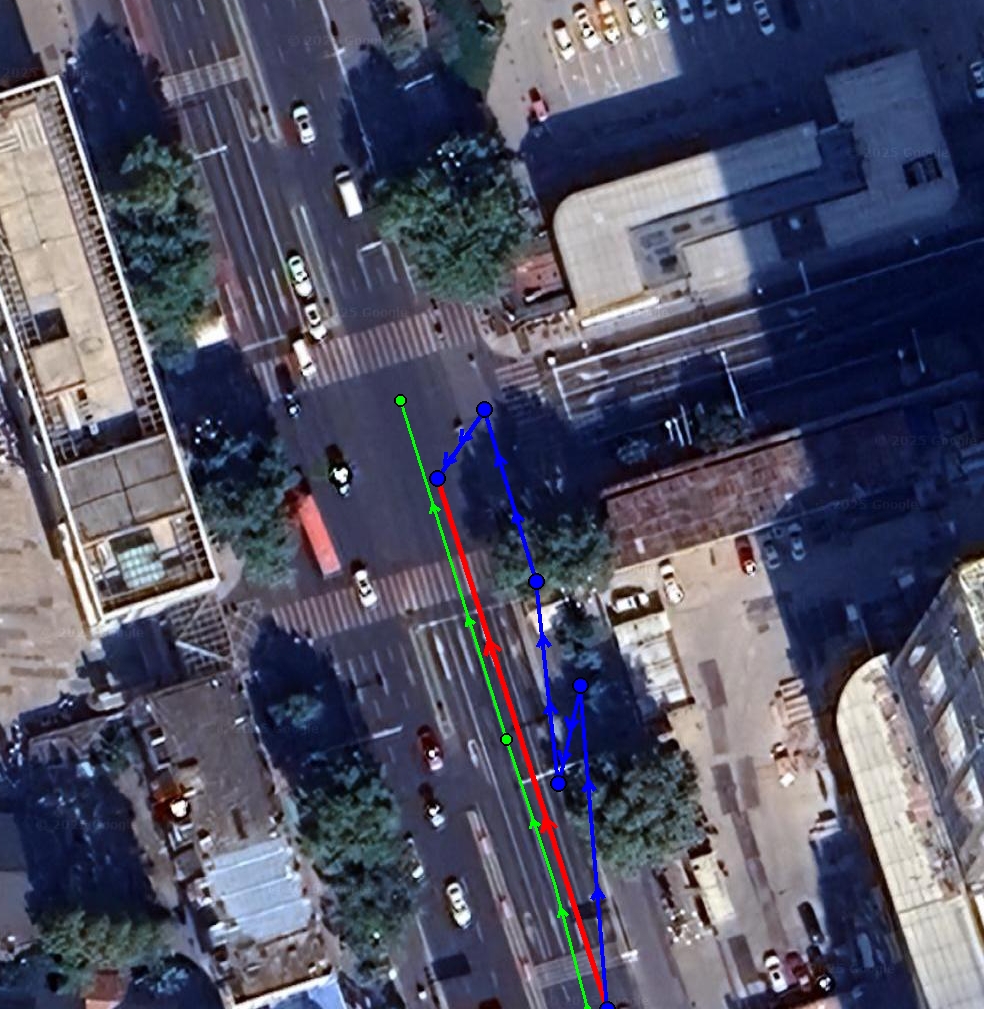}
    \caption{Case 2}
  \end{subfigure}
  \hfill
  \begin{subfigure}[b]{0.32\linewidth}
    \includegraphics[width=\linewidth]{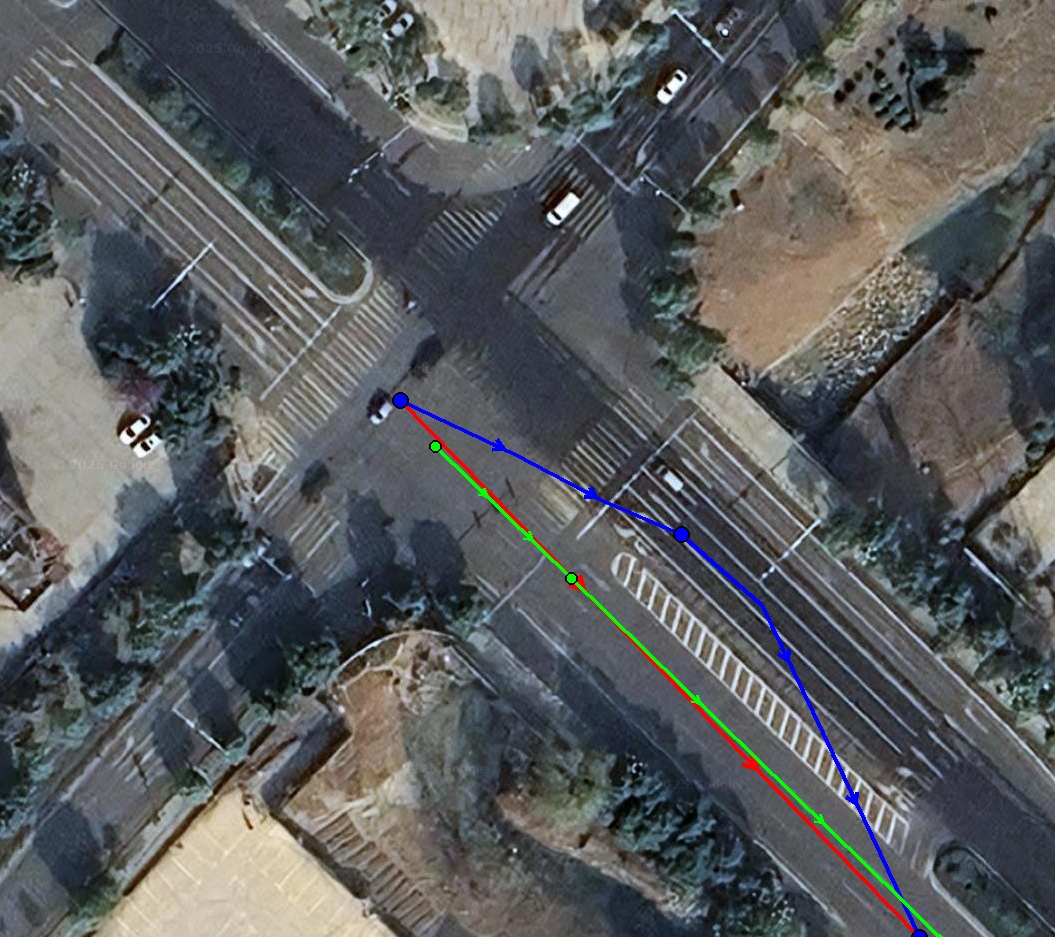}
    \caption{Case 3}
  \end{subfigure}
  \caption{Example images of BEV road network correction failure cases.}
  \label{fig:badcases:task3}
\end{figure}

\begin{enhancedpromptbox}

\textbf{Case 1:}

\textbf{MLLM Response:}

\begin{lstlisting}
```yaml
# Explanation: Identified two junctions where the road splits/intersects and created line segments connecting them based on the reference line's direction. Corrected the line segments to follow the actual road centerlines.
junctions: ["POINT (641 640)", "POINT (501 855)"]
lines: ["LINESTRING (840 400, 641 640)", "LINESTRING (641 640, 501 855)", "LINESTRING (501 855, 400 1062)"]
```
\end{lstlisting}

\textbf{Analysis:} The model cannot precisely determine the location of junctions, leading to inaccurate junction coordinates and incorrect road network topology.

\end{enhancedpromptbox}

\begin{enhancedpromptbox}

\textbf{Case 2:}

\textbf{MLLM Response:}

\begin{lstlisting}
```yaml
# Explanation: Added junctions at the intersection identified in the image and split the original line into multiple segments following the road centerline. Corrected the path to accurately reflect traffic flow.
junctions: ["POINT (536 581)", "POINT (484 409)", "POINT (437 478)", "POINT (607 1009)", "POINT (558 783)", "POINT (580 685)"]
lines: ["LINESTRING (607 1009, 580 685)", "LINESTRING (580 685, 558 783)", "LINESTRING (558 783, 536 581)", "LINESTRING (536 581, 484 409)", "LINESTRING (484 409, 437 478)"]
```
\end{lstlisting}

\textbf{Analysis:} The model cannot output stable coordinates, producing inconsistent and unreliable junction and line segment positions across different reasoning attempts.

\end{enhancedpromptbox}

\begin{enhancedpromptbox}

\textbf{Case 3:}

\textbf{MLLM Response:}

\begin{lstlisting}
```yaml
# Explanation: The reference line missed the intersection. I added a junction at the intersection and split the line into two segments accordingly.
junctions: ["POINT (400 400)", "POINT (681 534)", "POINT (919 937)"]
lines: ["LINESTRING (400 400, 515 454, 602 500, 681 534)", "LINESTRING (681 534, 761 604, 863 818, 919 937)"]
```
\end{lstlisting}

\textbf{Analysis:} While the model recognizes that a junction is missing, it cannot accurately determine the direction of bidirectional roads, leading to incorrect road network topology.

\end{enhancedpromptbox}

\textbf{Summary:} Road network correction failures expose significant challenges in spatial reasoning: (1) imprecise junction localization, where models struggle to accurately identify intersection locations, (2) coordinate instability, resulting in inconsistent outputs across different reasoning attempts, and (3) difficulty in understanding bidirectional road configurations, leading to incorrect direction assignments in the road network.

\subsection{FPV Lane Counting}

This subsection presents failure cases generated by GPT-5 in FPV lane counting tasks.

\begin{figure}[h]
  \centering
  \begin{subfigure}[b]{0.48\linewidth}
    \includegraphics[width=\linewidth]{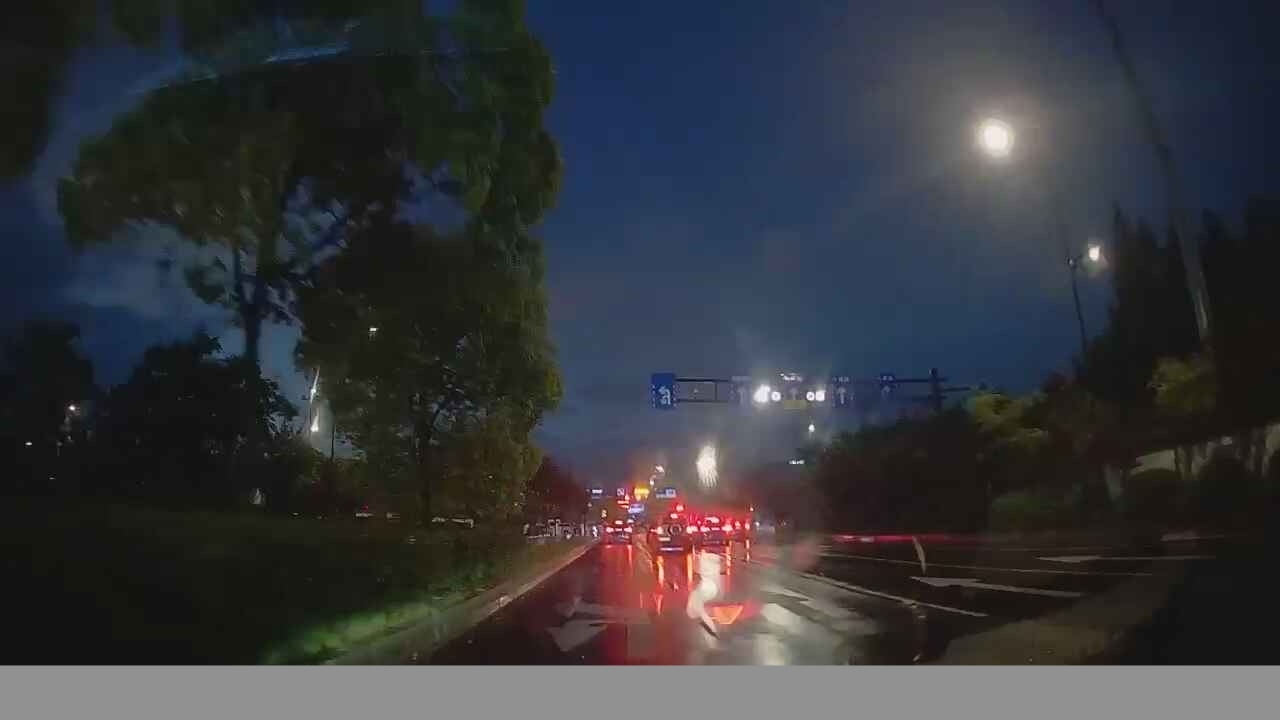}
    \caption{Case 1}
  \end{subfigure}
  \hfill
  \begin{subfigure}[b]{0.48\linewidth}
    \includegraphics[width=\linewidth]{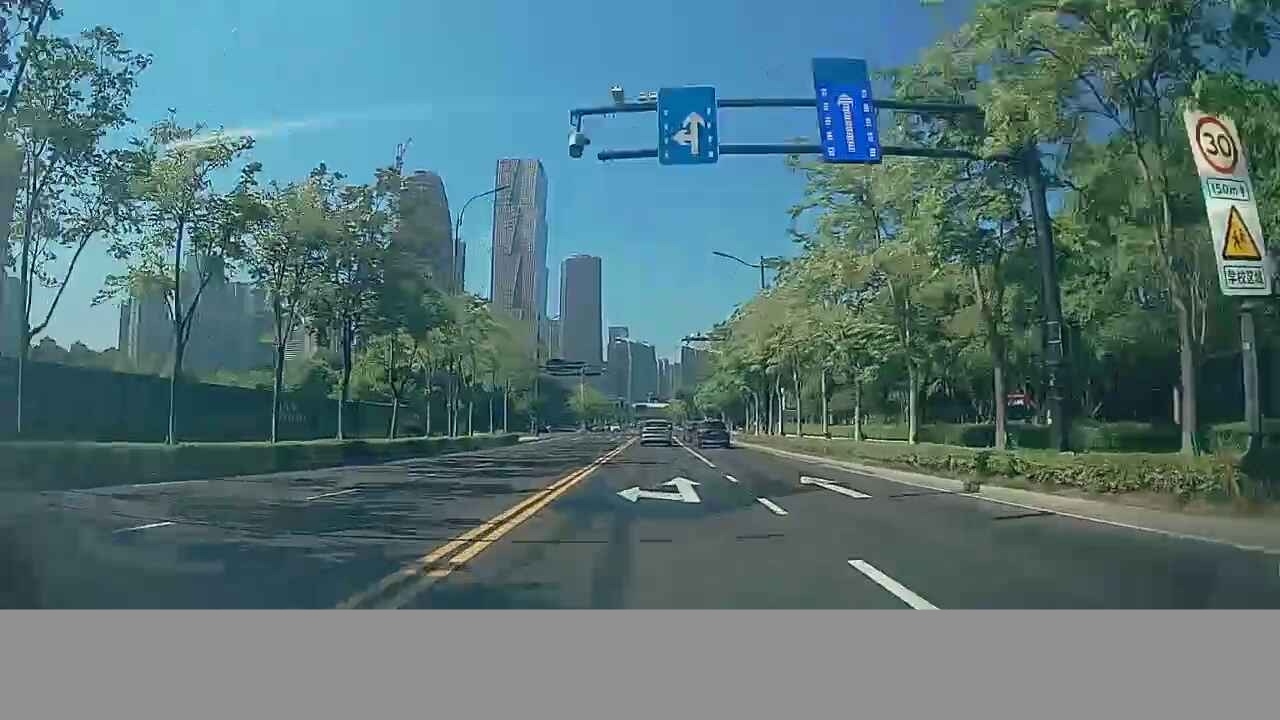}
    \caption{Case 2}
  \end{subfigure}
  \caption{Example images of FPV lane counting failure cases.}
  \label{fig:badcases:task4}
\end{figure}

\begin{enhancedpromptbox}

\textbf{Case 1:}

\textbf{MLLM Response:}

\begin{lstlisting}
```yaml
# Explanation: The wet road shows three parallel lanes with visible lane dividers and arrow markings, indicating three lanes moving in our direction.
num_lanes: 3
```
\end{lstlisting}

\textbf{Ground Truth:}

\begin{lstlisting}
```yaml
num_lanes: 5
```
\end{lstlisting}

\textbf{Analysis:} Under poor nighttime lighting conditions, the model cannot utilize overhead arrow signs to infer the number of lanes, relying solely on visible ground markings which are insufficient in low-light scenarios.

\end{enhancedpromptbox}

\begin{enhancedpromptbox}

\textbf{Case 2:}

\textbf{MLLM Response:}

\begin{lstlisting}
```yaml
# Explanation: Three distinct lanes are visible on our side-left next to the double yellow centerline, a middle lane, and a rightmost lane by the curb-separated by dashed white lines and lane arrows.
num_lanes: 3
```
\end{lstlisting}

\textbf{Ground Truth:}

\begin{lstlisting}
```yaml
num_lanes: 5
```
\end{lstlisting}

\textbf{Analysis:} The model cannot understand that opposite-direction lanes are unrelated to the current direction and should not be included in the lane count calculation.

\end{enhancedpromptbox}

\textbf{Summary:} In the FPV perspective, MLLMs are further exposed for their problems in urban roadway marking and spatial understanding, especially for misrecognition and understanding of opposing lanes. In addition, poor lighting conditions are one of the challenges for the application of MLLMs in realistic urban scenarios.

\subsection{FPV Lane Designation Recognition}

This subsection presents failure cases generated by GPT-5 in FPV lane designation recognition tasks.

\begin{figure}[h]
  \centering
  \begin{subfigure}[b]{0.48\linewidth}
    \includegraphics[width=\linewidth]{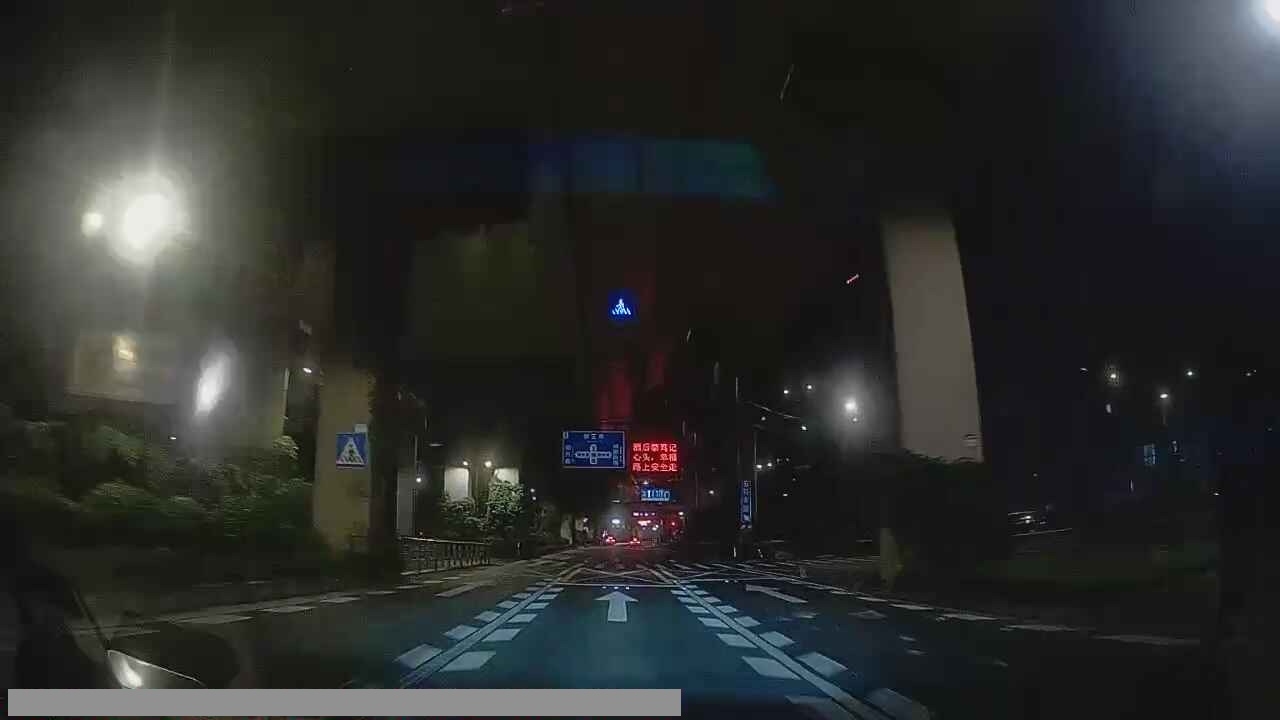}
    \caption{Case 1}
  \end{subfigure}
  \hfill
  \begin{subfigure}[b]{0.48\linewidth}
    \includegraphics[width=\linewidth]{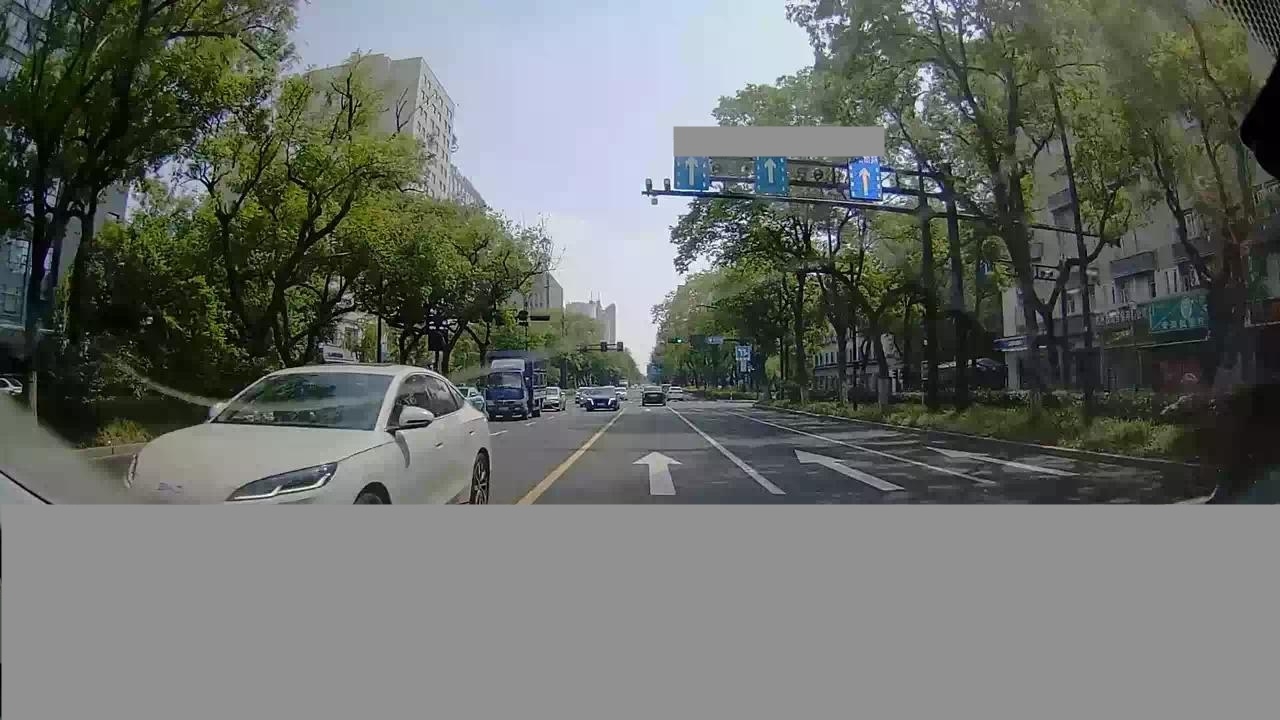}
    \caption{Case 2}
  \end{subfigure}
  \caption{Example images of FPV lane designation failure cases.}
  \label{fig:badcases:task5}
\end{figure}

\begin{enhancedpromptbox}

\textbf{Case 1:}

\textbf{MLLM Response:}

\begin{lstlisting}
```yaml
# Explanation: The center lane clearly shows a straight arrow marking, and typical Chinese intersection layout plus the overhead lane sign suggest the left lane is for left-turns and the right lane is for right-turns.
lane_designations: [["left-turn"], ["straight"], ["right-turn"]]
```
\end{lstlisting}

\textbf{Ground Truth:}

\begin{lstlisting}
```yaml
lane_designations: [["straight"], ["straight"], ["straight"]]
```
\end{lstlisting}

\textbf{Analysis:} The model relies excessively on common sense reasoning rather than visual information from the image, leading to incorrect lane function assignments based on typical intersection layouts rather than actual road markings.

\end{enhancedpromptbox}

\begin{enhancedpromptbox}

\textbf{Case 2:}

\textbf{MLLM Response:}

\begin{lstlisting}
```yaml
# Center lane clearly shows a straight arrow; rightmost lane marking shows straight and right-turn; leftmost lane has no visible turn markings, so assumed straight
lane_designations: [["straight"], ["straight"], ["straight", "right-turn"]]
```
\end{lstlisting}

\textbf{Ground Truth:}

\begin{lstlisting}
```yaml
lane_designations: [["straight"], ["straight"], ["straight"]]
```
\end{lstlisting}

\textbf{Analysis:} The model incorrectly identifies lanes from the opposite direction as lanes in the current direction, demonstrating confusion in bidirectional road scenarios.

\end{enhancedpromptbox}

\textbf{Summary:} Similarly, MLLMs are still not effective in this task to correctly understand lane direction and make correct inferences based on road markings as well as vehicle orientation. Also in nighttime environments, MLLMs seem to give up reasoning using road marking information that is recognizable to the human eye, and instead rely on common sense.

\subsection{FPV Road Type Classification}

This subsection presents failure cases generated by Gemini-2.5-Pro in FPV road type classification tasks.

\begin{figure}[h]
  \centering
  \begin{subfigure}[b]{0.48\linewidth}
    \includegraphics[width=\linewidth]{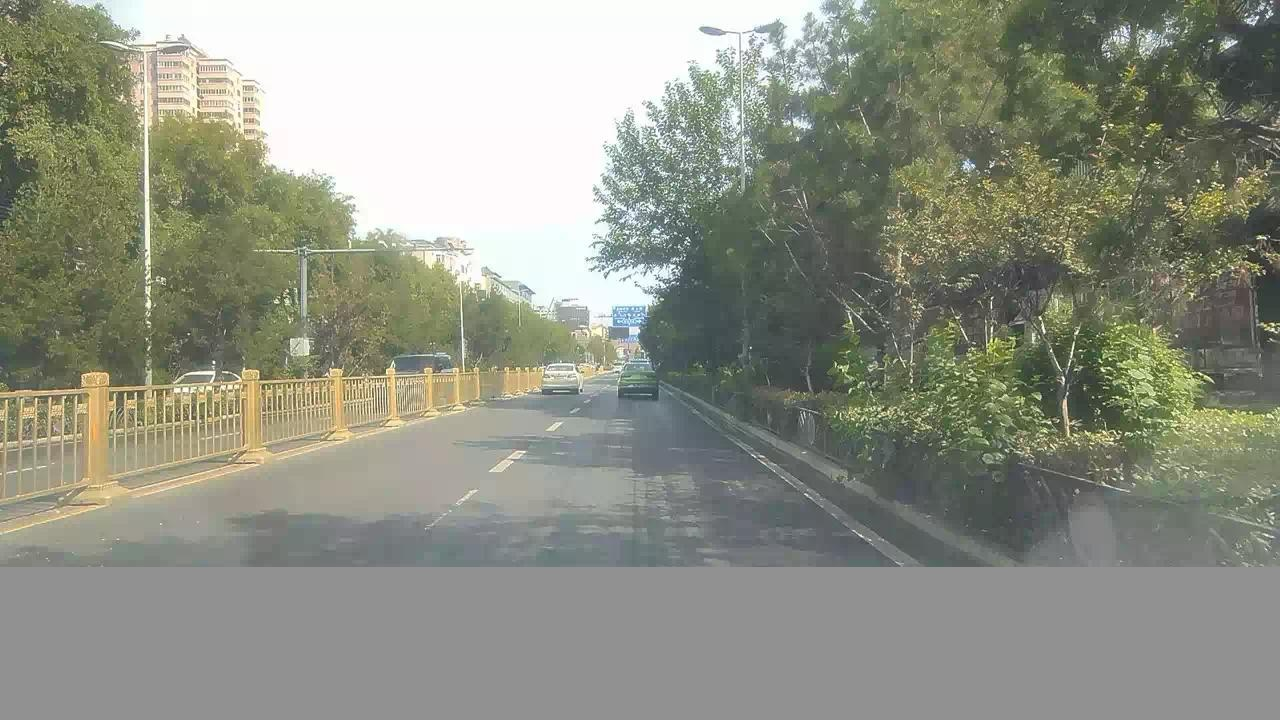}
    \caption{Case 1}
  \end{subfigure}
  \hfill
  \begin{subfigure}[b]{0.48\linewidth}
    \includegraphics[width=\linewidth]{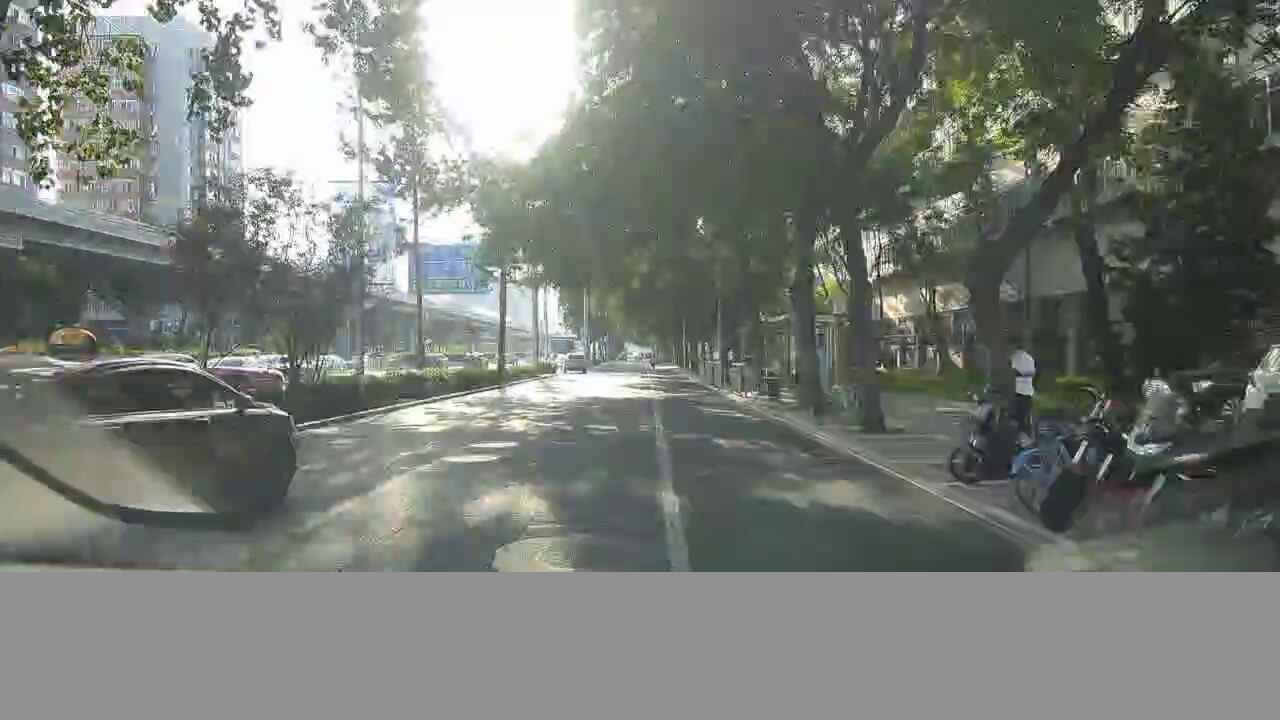}
    \caption{Case 2}
  \end{subfigure}
  \caption{Example images of FPV road type classification failure cases.}
  \label{fig:badcases:task6}
\end{figure}

\begin{enhancedpromptbox}

\textbf{Case 1:}

\textbf{MLLM Response:}

\begin{lstlisting}
```yaml
# Explanation: The vehicle is on a multi-lane road running parallel to another traffic system, separated by a yellow guardrail, which is a common design for a service road (auxiliary road) alongside a main road or expressway.
road_type: "service"
```
\end{lstlisting}

\textbf{Ground Truth:}

\begin{lstlisting}
```yaml
road_type: "main"
```
\end{lstlisting}

\textbf{Analysis:} The model cannot correctly identify that the central guardrail serves to separate bidirectional lanes rather than separate a service road from a main road, leading to incorrect road type classification.

\end{enhancedpromptbox}

\begin{enhancedpromptbox}

\textbf{Case 2:}

\textbf{MLLM Response:}

\begin{lstlisting}
```yaml
# Explanation: The image displays a wide, multi-lane road with an elevated highway to the left and a large overhead traffic sign in the distance, all indicative of a primary traffic artery rather than a smaller, auxiliary service road.
road_type: "main"
```
\end{lstlisting}

\textbf{Ground Truth:}

\begin{lstlisting}
```yaml
road_type: "service"
```
\end{lstlisting}

\textbf{Analysis:} The model incorrectly associates overhead signs and elevated structures in the distance with the current road, failing to recognize that these elements belong to a different road system rather than the current roadway.

\end{enhancedpromptbox}

\textbf{Summary:} Similarly, MLLMs still struggle with understanding lane direction. Meanwhile, the behavior of randomly associating visual elements hints at the risk of applying MLLMs in reality.

\bigskip

\subsection{Analysis}
\label{sec:diagnostic_analysis}

Beyond aggregate scores, we analyze representative failure cases to identify why current MLLMs struggle on RoadBench. The errors mainly fall into five aspects.
First, models often miss subtle visual cues, such as thin lane dividers, small turning arrows, worn markings, shadows, occlusions, and low-contrast road surfaces. This directly affects lane counting and lane designation recognition.
Second, models frequently fail to organize visible markings into correct spatial relations. Typical errors include confusing lane order, mixing target lanes with opposite-direction lanes, misjudging driving direction, and mismatching markings with the corresponding lane or centerline.
Third, models do not reliably combine visual evidence with traffic-domain knowledge. For example, they may recognize individual arrows but still infer incorrect lane functions in multi-lane, bidirectional, or complex junction scenes.
Fourth, cross-view tasks introduce additional difficulty because BEV and FPV images differ in viewpoint, scale, and visible context. Models therefore often fail to establish consistent scene-level correspondence across the two views.
Finally, in structured tasks such as road-network correction, models sometimes produce incomplete or schema-inconsistent outputs, indicating limited reliability in converting spatial reasoning into precise structured representations.
These observations show that RoadBench reveals not only performance gaps, but also concrete weaknesses in fine-grained perception, direction-aware spatial reasoning, domain-knowledge integration, cross-view consistency, and structured spatial prediction.

%% file: main.bbl
\begin{thebibliography}{10}
\providecommand{\url}[1]{\texttt{#1}}
\providecommand{\urlprefix}{URL }
\providecommand{\doi}[1]{https://doi.org/#1}

\bibitem{achiam2023gpt}
Achiam, J., Adler, S., Agarwal, S., Ahmad, L., Akkaya, I., Aleman, F.L., Almeida, D., Altenschmidt, J., Altman, S., Anadkat, S., et~al.: Gpt-4 technical report. arXiv preprint arXiv:2303.08774  (2023)

\bibitem{azzolini2025cosmos}
Azzolini, A., Bai, J., Brandon, H., Cao, J., Chattopadhyay, P., Chen, H., Chu, J., Cui, Y., Diamond, J., Ding, Y., et~al.: Cosmos-reason1: From physical common sense to embodied reasoning. arXiv preprint arXiv:2503.15558  (2025)

\bibitem{bai2025qwen3}
Bai, S., Cai, Y., Chen, R., Chen, K., Chen, X., Cheng, Z., Deng, L., Ding, W., Gao, C., Ge, C., et~al.: Qwen3-vl technical report. arXiv preprint arXiv:2511.21631  (2025)

\bibitem{bai2025qwen2}
Bai, S., Chen, K., Liu, X., Wang, J., Ge, W., Song, S., Dang, K., Wang, P., Wang, S., Tang, J., et~al.: Qwen2. 5-vl technical report. arXiv preprint arXiv:2502.13923  (2025)

\bibitem{brown2020language}
Brown, T., Mann, B., Ryder, N., Subbiah, M., Kaplan, J.D., Dhariwal, P., Neelakantan, A., Shyam, P., Sastry, G., Askell, A., et~al.: Language models are few-shot learners. Advances in neural information processing systems  \textbf{33},  1877--1901 (2020)

\bibitem{chang2025driving}
Chang, X., Xue, M., Liu, X., Pan, Z., Wei, X.: Driving by the rules: A benchmark for integrating traffic sign regulations into vectorized hd map. In: Proceedings of the Computer Vision and Pattern Recognition Conference. pp. 6823--6833 (2025)

\bibitem{chen2024spatialvlm}
Chen, B., Xu, Z., Kirmani, S., Ichter, B., Sadigh, D., Guibas, L., Xia, F.: Spatialvlm: Endowing vision-language models with spatial reasoning capabilities. In: Proceedings of the IEEE/CVF Conference on Computer Vision and Pattern Recognition (CVPR). pp. 14455--14465 (June 2024)

\bibitem{chen2025spatialllm}
Chen, J., Wang, H., Li, J., Liu, Y., Dong, Z., Yang, B.: Spatialllm: From multi-modality data to urban spatial intelligence. arXiv preprint arXiv:2505.12703  (2025)

\bibitem{cheng2024spatialrgpt}
Cheng, A.C., Yin, H., Fu, Y., Guo, Q., Yang, R., Kautz, J., Wang, X., Liu, S.: Spatialrgpt: Grounded spatial reasoning in vision-language models. Advances in Neural Information Processing Systems  \textbf{37},  135062--135093 (2024)

\bibitem{comanici2025gemini}
Comanici, G., Bieber, E., Schaekermann, M., Pasupat, I., Sachdeva, N., Dhillon, I., Blistein, M., Ram, O., Zhang, D., Rosen, E., et~al.: Gemini 2.5: Pushing the frontier with advanced reasoning, multimodality, long context, and next generation agentic capabilities. arXiv preprint arXiv:2507.06261  (2025)

\bibitem{cui2024survey}
Cui, C., Ma, Y., Cao, X., Ye, W., Zhou, Y., Liang, K., Chen, J., Lu, J., Yang, Z., Liao, K.D., et~al.: A survey on multimodal large language models for autonomous driving. In: Proceedings of the IEEE/CVF winter conference on applications of computer vision. pp. 958--979 (2024)

\bibitem{daxberger2025mm}
Daxberger, E., Wenzel, N., Griffiths, D., Gang, H., Lazarow, J., Kohavi, G., Kang, K., Eichner, M., Yang, Y., Dehghan, A., et~al.: Mm-spatial: Exploring 3d spatial understanding in multimodal llms. In: Proceedings of the IEEE/CVF International Conference on Computer Vision. pp. 7395--7408 (2025)

\bibitem{driess2023palm}
Driess, D., Xia, F., Sajjadi, M.S., Lynch, C., Chowdhery, A., Ichter, B., Wahid, A., Tompson, J., Vuong, Q., Yu, T., et~al.: Palm-e: An embodied multimodal language model. In: International Conference on Machine Learning. pp. 8469--8488. PMLR (2023)

\bibitem{eiter1994computing}
Eiter, T., Mannila, H., et~al.: Computing discrete fr{\'e}chet distance  (1994)

\bibitem{feng2025citygpt}
Feng, J., Liu, T., Du, Y., Guo, S., Lin, Y., Li, Y.: Citygpt: Empowering urban spatial cognition of large language models. In: Proceedings of the 31st ACM SIGKDD Conference on Knowledge Discovery and Data Mining V. 2. pp. 591--602 (2025)

\bibitem{feng2025urbanllava}
Feng, J., Wang, S., Liu, T., Xi, Y., Li, Y.: Urbanllava: A multi-modal large language model for urban intelligence with spatial reasoning and understanding. In: Proceedings of the IEEE/CVF International Conference on Computer Vision. pp. 6209--6219 (2025)

\bibitem{feng2025citybench}
Feng, J., Zhang, J., Liu, T., Zhang, X., Ouyang, T., Yan, J., Du, Y., Guo, S., Li, Y.: Citybench: Evaluating the capabilities of large language models for urban tasks. In: Proceedings of the 31st ACM SIGKDD Conference on Knowledge Discovery and Data Mining V. 2. pp. 5413--5424 (2025)

\bibitem{fu2024mmecomprehensiveevaluationbenchmark}
Fu, C., Chen, P., Shen, Y., Qin, Y., Zhang, M., Lin, X., Yang, J., Zheng, X., Li, K., Sun, X., Wu, Y., Ji, R.: Mme: A comprehensive evaluation benchmark for multimodal large language models (2024), \url{https://arxiv.org/abs/2306.13394}

\bibitem{guo2024regiongpt}
Guo, Q., De~Mello, S., Yin, H., Byeon, W., Cheung, K.C., Yu, Y., Luo, P., Liu, S.: Regiongpt: Towards region understanding vision language model. In: Proceedings of the IEEE/CVF Conference on Computer Vision and Pattern Recognition. pp. 13796--13806 (2024)

\bibitem{hu2025rs}
Hu, H., Wang, P., Bi, H., Tong, B., Wang, Z., Diao, W., Chang, H., Feng, Y., Zhang, Z., Wang, Y., et~al.: Rs-vheat: Heat conduction guided efficient remote sensing foundation model. In: Proceedings of the IEEE/CVF International Conference on Computer Vision. pp. 9876--9887 (2025)

\bibitem{li2024seed}
Li, B., Ge, Y., Ge, Y., Wang, G., Wang, R., Zhang, R., Shan, Y.: Seed-bench: Benchmarking multimodal large language models. In: Proceedings of the IEEE/CVF Conference on Computer Vision and Pattern Recognition. pp. 13299--13308 (2024)

\bibitem{li2022blip}
Li, J., Li, D., Xiong, C., Hoi, S.: Blip: Bootstrapping language-image pre-training for unified vision-language understanding and generation. In: International conference on machine learning. pp. 12888--12900. PMLR (2022)

\bibitem{li2024manipllm}
Li, X., Zhang, M., Geng, Y., Geng, H., Long, Y., Shen, Y., Zhang, R., Liu, J., Dong, H.: Manipllm: Embodied multimodal large language model for object-centric robotic manipulation. In: Proceedings of the IEEE/CVF Conference on Computer Vision and Pattern Recognition. pp. 18061--18070 (2024)

\bibitem{liu2024mmbench}
Liu, Y., Duan, H., Zhang, Y., Li, B., Zhang, S., Zhao, W., Yuan, Y., Wang, J., He, C., Liu, Z., et~al.: Mmbench: Is your multi-modal model an all-around player? In: European conference on computer vision. pp. 216--233. Springer (2024)

\bibitem{lu2023mathvista}
Lu, P., Bansal, H., Xia, T., Liu, J., Li, C., Hajishirzi, H., Cheng, H., Chang, K.W., Galley, M., Gao, J.: Mathvista: Evaluating mathematical reasoning of foundation models in visual contexts. arXiv preprint arXiv:2310.02255  (2023)

\bibitem{labelu_2025}
OpenDataLab: Labelu (2025), \url{https://github.com/opendatalab/labelU-kit}, accessed: 2025-09-01

\bibitem{ou2025geopix}
Ou, R., Hu, Y., Zhang, F., Chen, J., Liu, Y.: Geopix: A multimodal large language model for pixel-level image understanding in remote sensing. IEEE Geoscience and Remote Sensing Magazine  (2025)

\bibitem{park2025nuplanqa}
Park, S.Y., Cui, C., Ma, Y., Moradipari, A., Gupta, R., Han, K., Wang, Z.: Nuplanqa: A large-scale dataset and benchmark for multi-view driving scene understanding in multi-modal large language models. arXiv preprint arXiv:2503.12772  (2025)

\bibitem{radford2021learning}
Radford, A., Kim, J.W., Hallacy, C., Ramesh, A., Goh, G., Agarwal, S., Sastry, G., Askell, A., Mishkin, P., Clark, J., et~al.: Learning transferable visual models from natural language supervision. In: International conference on machine learning. pp. 8748--8763. PMLR (2021)

\bibitem{ramakrishnan2024does}
Ramakrishnan, S.K., Wijmans, E., Kraehenbuehl, P., Koltun, V.: Does spatial cognition emerge in frontier models? arXiv preprint arXiv:2410.06468  (2024)

\bibitem{roberts2024charting}
Roberts, J., L{\"u}ddecke, T., Sheikh, R., Han, K., Albanie, S.: Charting new territories: Exploring the geographic and geospatial capabilities of multimodal llms. In: Proceedings of the IEEE/CVF Conference on Computer Vision and Pattern Recognition. pp. 554--563 (2024)

\bibitem{team2025gemma}
Team, G., Kamath, A., Ferret, J., Pathak, S., Vieillard, N., Merhej, R., Perrin, S., Matejovicova, T., Ram{\'e}, A., Rivi{\`e}re, M., et~al.: Gemma 3 technical report. arXiv preprint arXiv:2503.19786  (2025)

\bibitem{vteam2025glm45vglm41vthinkingversatilemultimodal}
Team, V., Hong, W., Yu, W., Gu, X., Wang, G., Gan, G., Tang, H., Cheng, J., Qi, J., Ji, J., Pan, L., Duan, S., Wang, W., Wang, Y., Cheng, Y., He, Z., Su, Z., Yang, Z., Pan, Z., Zeng, A., Wang, B., Chen, B., Shi, B., Pang, C., Zhang, C., Yin, D., Yang, F., Chen, G., Xu, J., Zhu, J., Chen, J., Chen, J., Chen, J., Lin, J., Wang, J., Chen, J., Lei, L., Gong, L., Pan, L., Liu, M., Xu, M., Zhang, M., Zheng, Q., Yang, S., Zhong, S., Huang, S., Zhao, S., Xue, S., Tu, S., Meng, S., Zhang, T., Luo, T., Hao, T., Tong, T., Li, W., Jia, W., Liu, X., Zhang, X., Lyu, X., Fan, X., Huang, X., Wang, Y., Xue, Y., Wang, Y., Wang, Y., An, Y., Du, Y., Shi, Y., Huang, Y., Niu, Y., Wang, Y., Yue, Y., Li, Y., Zhang, Y., Wang, Y., Wang, Y., Zhang, Y., Xue, Z., Hou, Z., Du, Z., Wang, Z., Zhang, P., Liu, D., Xu, B., Li, J., Huang, M., Dong, Y., Tang, J.: Glm-4.5v and glm-4.1v-thinking: Towards versatile multimodal reasoning with scalable reinforcement learning (2025), \url{https://arxiv.org/abs/2507.01006}

\bibitem{tian2025drivevlm}
Tian, X., Gu, J., Li, B., Liu, Y., Wang, Y., Zhao, Z., Zhan, K., Jia, P., Lang, X., Zhao, H.: Drivevlm: The convergence of autonomous driving and large vision-language models. In: Conference on Robot Learning. pp. 4698--4726. PMLR (2025)

\bibitem{touvron2023llama}
Touvron, H., Lavril, T., Izacard, G., Martinet, X., Lachaux, M.A., Lacroix, T., Rozi{\`e}re, B., Goyal, N., Hambro, E., Azhar, F., et~al.: Llama: Open and efficient foundation language models. arXiv preprint arXiv:2302.13971  (2023)

\bibitem{wang2024picture}
Wang, J., Ming, Y., Shi, Z., Vineet, V., Wang, X., Li, S., Joshi, N.: Is a picture worth a thousand words? delving into spatial reasoning for vision language models. Advances in Neural Information Processing Systems  \textbf{37},  75392--75421 (2024)

\bibitem{wang2025spatial457}
Wang, X., Ma, W., Zhang, T., de~Melo, C.M., Chen, J., Yuille, A.: Spatial457: A diagnostic benchmark for 6d spatial reasoning of large multimodal models. In: Proceedings of the Computer Vision and Pattern Recognition Conference. pp. 24669--24679 (2025)

\bibitem{wei2024slow}
Wei, H., Yin, Y., Li, Y., Wang, J., Zhao, L., Sun, J., Ge, Z., Zhang, X., Jiang, D.: Slow perception: Let's perceive geometric figures step-by-step. arXiv preprint arXiv:2412.20631  (2024)

\bibitem{xiao2024comprehensive}
Xiao, H., Zhou, F., Liu, X., Liu, T., Li, Z., Liu, X., Huang, X.: A comprehensive survey of large language models and multimodal large language models in medicine. Information Fusion p. 102888 (2024)

\bibitem{xie2025vlms}
Xie, S., Kong, L., Dong, Y., Sima, C., Zhang, W., Chen, Q.A., Liu, Z., Pan, L.: Are vlms ready for autonomous driving? an empirical study from the reliability, data, and metric perspectives. arXiv preprint arXiv:2501.04003  (2025)

\bibitem{xuan2025dynamicvl}
Xuan, W., Wang, J., Qi, H., Chen, Z., Zheng, Z., Zhong, Y., Xia, J., Yokoya, N.: Dynamicvl: Benchmarking multimodal large language models for dynamic city understanding. arXiv preprint arXiv:2505.21076  (2025)

\bibitem{yang2025thinking}
Yang, J., Yang, S., Gupta, A.W., Han, R., Fei-Fei, L., Xie, S.: Thinking in space: How multimodal large language models see, remember, and recall spaces. In: Proceedings of the Computer Vision and Pattern Recognition Conference. pp. 10632--10643 (2025)

\bibitem{yin2024survey}
Yin, S., Fu, C., Zhao, S., Li, K., Sun, X., Xu, T., Chen, E.: A survey on multimodal large language models. National Science Review  \textbf{11}(12),  nwae403 (2024)

\bibitem{zhang2025vldrive}
Zhang, R., Zhang, W., Tan, X., Yang, S., Wan, X., Luo, X., Li, G.: Vldrive: Vision-augmented lightweight mllms for efficient language-grounded autonomous driving. In: Proceedings of the IEEE/CVF International Conference on Computer Vision. pp. 5923--5933 (2025)

\bibitem{zhang2025open}
Zhang, S., Chen, A., Sun, Y., Gu, J., Zheng, Y.Y., Koniusz, P., Zou, K., Hengel, A.v.d., Xue, Y.: Open eyes, then reason: Fine-grained visual mathematical understanding in mllms. arXiv preprint arXiv:2501.06430  (2025)

\bibitem{zhang2024earthmarker}
Zhang, W., Cai, M., Zhang, T., Zhuang, Y., Li, J., Mao, X.: Earthmarker: A visual prompting multi-modal large language model for remote sensing. IEEE Transactions on Geoscience and Remote Sensing  (2024)

\bibitem{zhang2026urbanmllm}
Zhang, X., Ouyang, T., Shang, Y., Liao, Q., Li, Y.: Urban{MLLM}: Joint learning of cross-view imagery for urban understanding. In: Forty-third International Conference on Machine Learning (2026), \url{https://openreview.net/forum?id=Av10FtirO4}

\bibitem{zhang2024interndrive}
Zhang, Y., Nie, Y.: Interndrive: A multimodal large language model for autonomous driving scenario understanding. In: Proceedings of the 2024 4th International Conference on Artificial Intelligence, Automation and High Performance Computing. pp. 294--305 (2024)

\bibitem{zhao2025embodied}
Zhao, B., Wang, Z., Fang, J., Gao, C., Man, F., Cui, J., Wang, X., Chen, X., Li, Y., Zhu, W.: Embodied-r: Collaborative framework for activating embodied spatial reasoning in foundation models via reinforcement learning. In: Proceedings of the 33rd ACM International Conference on Multimedia. pp. 11071--11080 (2025)

\bibitem{zhao2025cityeqa}
Zhao, Y., Xu, K., Zhu, Z., Hu, Y., Zheng, Z., Chen, Y., Ji, Y., Gao, C., Li, Y., Huang, J.: Cityeqa: A hierarchical llm agent on embodied question answering benchmark in city space. EMNLP  (2025)

\bibitem{zhou2025urbench}
Zhou, B., Yang, H., Chen, D., Ye, J., Bai, T., Yu, J., Zhang, S., Lin, D., He, C., Li, W.: Urbench: A comprehensive benchmark for evaluating large multimodal models in multi-view urban scenarios. In: Proceedings of the AAAI Conference on Artificial Intelligence. vol.~39, pp. 10707--10715 (2025)

\end{thebibliography}
